\documentclass[11pt]{article}

\usepackage[margin=1in]{geometry}
\usepackage[T1]{fontenc}
\usepackage{lmodern}

\usepackage{graphicx}
\usepackage{multirow}
\usepackage{amsmath,amssymb,amsfonts}
\usepackage{amsthm}
\usepackage{mathrsfs}
\usepackage{xcolor}
\usepackage{textcomp}
\usepackage{booktabs}
\usepackage{algorithm}
\usepackage{algpseudocode}
\usepackage{listings}
\usepackage{array}
\usepackage{tabularx}
\usepackage{url}
\usepackage{float}
\usepackage{xurl}
\usepackage{grffile}
\usepackage{caption}
\usepackage{placeins}
\usepackage{dblfloatfix}

\usepackage[numbers,sort&compress]{natbib}


\usepackage{tikz}
\usepackage{pgfplots}
\pgfplotsset{compat=1.18}

\graphicspath{{figures/}}

\newtheorem{remark}{Remark}

\title{Stateful Guardrails for Multi-Turn LLM Systems: A Conversational Risk Accumulation Framework}
\author{%
  Sanjay Mishra\\Independent Researcher, Raleigh, NC 27601, USA\\\texttt{sanmish4@icloud.com}
  \and
  Divya Chukkapalli\\Independent Researcher, Apex, NC 27502, USA\\\texttt{divya.95j@gmail.com}
  \and
  Ganesh R.~Naik\\College of Medicine and Public Health, Flinders University, Bedford Park, SA 5042, Australia\\\texttt{ganesh.naik@flinders.edu.au}
}
\date{}

\begin{document}
\maketitle

\begin{abstract}
Most safety guardrails for large language models (LLMs) evaluate each prompt--response pair in isolation, which misses failures that arise only over a dialogue as benign turns compose into harm. We term this \emph{Conversational Risk Accumulation} (CRA): gradual intent drift, fragmented assembly of prohibited instructions, and sensitivity build-up from repeated disclosures.

We propose a session-layer CRA Framework that tracks three trajectory signals: semantic drift from a session anchor, a sensitivity-weighted information accumulation graph over extracted entities, and a compliance-gradient signal capturing increasing willingness to comply. For scoring, we provide (i)~an unsupervised convex fusion for attribution and ablations, and (ii)~\textbf{CRA-Net DA}, a compact learned trajectory model trained with family-adversarial objectives to reduce length and topic-coverage confounds.

To benchmark CRA, we release \textbf{CRA-Bench v0.1} (1{,}200 eight-turn sessions across three threat families with topic-matched benign twins), \textbf{CRA-Bench v0.2} (LLM-paraphrased variants to reduce template artifacts), and an extended \textbf{5-family} set (2{,}000 sessions adding persona priming and context stuffing). We introduce a trajectory-native evaluation protocol with session-level splits, mixed-set threshold calibration, Trajectory AUROC, turns-to-detection, calibrated false-positive metrics, bootstrap confidence intervals, leave-one-family-out diagnostic stress tests, and synthetic$\to$human transfer checks. Claims focus on within-distribution session scoring on CRA-Bench and human-transfer subsets.
\end{abstract}


\section{Introduction}\label{sec1}
LLMs are increasingly deployed in interactive settings (enterprise assistants,
retrieval-augmented generation (RAG) pipelines, and tool-using agents) where
users engage in extended, goal-directed dialogues that can span dozens of turns.
Most safety mechanisms in these systems remain \emph{turn-level}: each
prompt--response pair is evaluated in isolation, and the conversation proceeds
whenever no single exchange violates policy.

That design has a structural blind spot. A multi-turn session is not merely a
sequence of independent decisions; risk can emerge \emph{compositionally} across
turns. A user may fragment a prohibited request into individually benign
sub-queries, gradually condition the model into higher compliance through
incremental boundary erosion, or accumulate innocuous disclosures that together
constitute a privacy violation. In every such case, a turn-level filter approves
each message while the session quietly crosses into unsafe territory. The danger
is not in any single exchange. It is in the path.

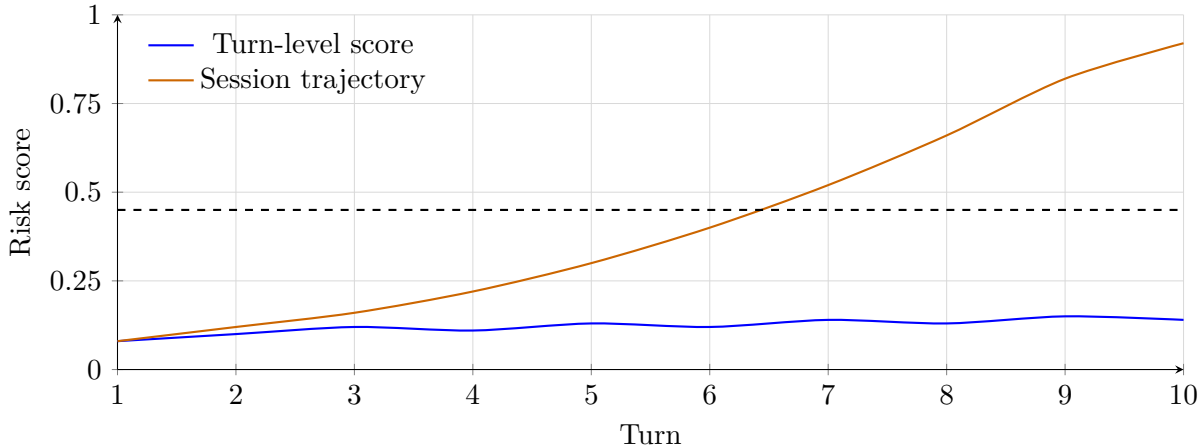
\begin{figure}[!t]
  \centering
  \begin{tikzpicture}
    \begin{axis}[
      width=0.95\linewidth,
      height=0.38\linewidth,
      xmin=1,xmax=10,
      ymin=0,ymax=1,
      axis lines=left,
      xlabel={Turn},
      ylabel={Risk score},
      ytick={0,0.25,0.5,0.75,1},
      legend style={at={(0.02,0.98)},anchor=north west,draw=none,fill=none},
      grid=both,
      major grid style={line width=.2pt,draw=gray!30},
      minor grid style={line width=.1pt,draw=gray!15},
    ]
      \addplot[smooth,thick,blue] coordinates {(1,0.08) (2,0.10) (3,0.12) (4,0.11) (5,0.13) (6,0.12) (7,0.14) (8,0.13) (9,0.15) (10,0.14)};
      \addlegendentry{Turn-level score}

      \addplot[smooth,thick,orange!80!black] coordinates {(1,0.08) (2,0.12) (3,0.16) (4,0.22) (5,0.30) (6,0.40) (7,0.52) (8,0.66) (9,0.82) (10,0.92)};
      \addlegendentry{Session trajectory}

      \addplot[dashed,thick,black] coordinates {(1,0.45) (10,0.45)};
      \node[anchor=west] at (axis cs:10,0.45) {\scriptsize $\tau$ (per-turn threshold)};
    \end{axis}
  \end{tikzpicture}
  \caption{Illustration of the CRA failure mode: turn-level risk scores can remain below
  a per-turn threshold while the session-level trajectory drifts into an unsafe region.
  CRA targets this gap by scoring the evolving conversation state rather than isolated turns.}
  \label{fig:cra_illustration}
\end{figure}

We call this failure class \emph{Conversational Risk Accumulation} (CRA): a
conversation becomes unsafe not because of one clearly disallowed turn, but
because of how multiple ``acceptable'' turns interact over time.
Figure~\ref{fig:cra_illustration} shows the gap between per-turn scores and the
underlying session trajectory.

\paragraph{Research aim.}
We pursue two goals: (i)~formalize CRA as a measurable multi-turn failure class,
and (ii)~propose and evaluate a session-layer scoring architecture that
aggregates cross-turn signals into a bounded, actionable intervention trigger.

\paragraph{Empirical motivation.}
Our \textbf{primary} evaluation uses CRA-Bench v0.1: CRA-positive and benign-twin sessions share the same user-turn count ($T{=}8$ in v0.1), so ranking metrics are not trivially explained by session length. We train CRA-Net only on the training split, calibrate the intervention threshold $\theta$ on a \emph{mixed} validation split, and report all headline numbers on a held-out test split from the \emph{same} benchmark distribution (Section~\ref{sec:primary-eval}).

CoSafe~\citep{cosafe2024} remains valuable at scale (1{,}800 sessions) but is \textbf{not} the primary evidence source: every CRA-positive session has three user turns and every CRA-negative session has one, so length-only classification is perfect (AUROC $=1.0$). We therefore treat CoSafe as a diagnostic stress test (Section~\ref{sec:cosafe-diagnostic}) and report length-matched and benign-traffic complements separately.

\paragraph{Contributions.}
This paper makes the following contributions:

\begin{itemize}
\item \textbf{Operational definition and threat taxonomy.} We define Conversational Risk Accumulation (CRA) in terms of measurable quantities on labelled multi-turn transcripts and introduce a five-type threat taxonomy that distinguishes accumulation mechanisms (Section~\ref{sec:taxonomy}).

\item \textbf{Three trajectory-level signals.} We instrument a session with a Semantic Drift Monitor, an Information Accumulation Graph with NIST-aligned sensitivity tiers~\citep{nist80060}, and a Compliance Gradient Detector, each defined as a per-turn function over the session history (Section~\ref{sec:framework}).

\item \textbf{CRA-Net DA as the primary fusion architecture.} We present family-adversarial CRA-Net (Section~\ref{sec:cranet-da}) as the deployable session scorer and retain convex fusion (Section~\ref{sec:convex_fusion}) as an interpretable diagnostic decomposition tool, not a competitive detector on CRA-Bench v0.2 (AUROC $\approx 0.60$ overall).

\item \textbf{CRA-Bench v0.1 and primary evaluation protocol.} We release a same-length benchmark (JSONL transcripts, onset labels, fixed seeds) and a session-level train/val/test protocol with mixed-split threshold calibration, benign FPR at scale, evasion variants, and baselines including CoSafe-native convex scoring, judge-LLM (published prompt), and CRA-Net ablations (no GRL; no $S_2/S_3$; feature-only MLP) (Sections~\ref{sec:cra_bench}, \ref{sec:primary-eval}).

\item \textbf{Synthetic-to-real transfer validation.} We release \textbf{Human-CRA-Transfer} ($N{=}972$: 750 human/red-team CoSafe gradual-escalation positives plus 222 length-matched ShareGPT benign negatives, all with three user turns) and show that CRA-Net trained \emph{only} on CRA-Bench v0.2 reaches AUROC $0.919$ on this held-out human corpus---closing the \emph{synthetic-to-real} gap (human-authored attacks never seen in training; Section~\ref{sec:human-transfer}).

\item \textbf{Diagnostic evaluation on CoSafe.} We report CoSafe under its known length confound and length-matched cross-corpus checks as \emph{complements}, not headline claims (Section~\ref{sec:cosafe-diagnostic}). An exploratory WildChat probe ($N{=}59$ per class) is relegated to a footnote because the sample is too small for reliable transfer statistics.

\item \textbf{Deployment patterns (illustrative).} We describe how the session layer \emph{can compose} with turn-level guardrails in enterprise RAG, agentic, and educational settings (Section~\ref{sec:deployment}); benign-FPR estimates use ShareGPT and public-chat proxies, not customer production logs.
\end{itemize}

\FloatBarrier
\section{Related Work}
\label{sec:related}

LLM safety research has primarily focused on turn-level behavior: harmful-content
suppression, refusal robustness, and policy compliance measured on individual
prompt--response pairs. HarmBench, for example, evaluates refusal under automated
red-teaming, but the dominant evaluation unit remains the single turn rather than
the full dialogue trajectory~\citep{rotger2024harmbench}.

A second major thread studies instruction override through jailbreaks and prompt
injection. Work on transferable jailbreak prompts catalogs a wide range of evasion
strategies designed to elicit disallowed content within a single
interaction~\citep{zou2024universal, wallace2019triggers, shen2024anything,
wei2023jailbroken, perez2022ignore}. Indirect prompt injection extends this
picture, showing that untrusted retrieved or embedded text can silently redirect
downstream behavior in LLM-integrated applications~\citep{greshake2023not}. These
findings directly motivate our Context Poisoning CRA type. What this literature
does not do,
however, is model how risk compounds across turns; success is still measured as a
per-turn breach.

\subsection{Multi-turn safety evaluation and dialogue state modeling}
A growing practical challenge is that real deployments are \emph{sessional}:
safety-relevant context is distributed across turns, and both attacks and
mitigations exploit conversation history. Yet much of the current evaluation
stack remains anchored in turn-level red-teaming suites and single-turn jailbreak
success criteria (e.g., HarmBench and universal jailbreak prompts)~\citep{rotger2024harmbench,
zou2024universal, wallace2019triggers}. This gap mirrors a classic distinction
in dialogue systems between tracking a latent \emph{dialogue state} across turns
(DST) and scoring individual utterances in isolation. CRA overlaps with DST in
that it maintains a compact, evolving session representation; however, it
departs in purpose and structure: rather than tracking task slots or belief
states, CRA tracks \emph{risk-relevant} state variables (semantic drift,
accumulated disclosures, refusal softening) and evaluates detectors with
trajectory-native metrics (e.g., time-to-detect) at an operating threshold.

\subsection{Multi-turn guardrails and trajectory classifiers}
Beyond benchmarks, several \emph{deployed} moderation systems now consume
conversation context rather than isolated turns. \citet{inan2023llamaguard}
introduced Llama Guard, an instruction-tuned moderator that classifies user
prompts and assistant responses within a multi-turn chat template;
\textbf{Llama Guard~3-1B} is the canonical Meta baseline but is gated on
HuggingFace, so we evaluate it via the public \texttt{llama-guard3:1b} Ollama
snapshot (reproducible without Meta approval). Follow-on systems such as
\textbf{Qwen3Guard} provide open-weight generative and streaming variants aimed
at long-context moderation~\citep{qwen2025guard} and serve as our primary
reproducible guardrail stand-in when gated weights are unavailable.
\citet{yu2024cosafe} study multi-turn \emph{coreference} attacks and release a
1{,}400-prompt dataset; the CoSafe rollouts corpus we use diagnostically
(Section~\ref{sec:cosafe-diagnostic}) extends this line but exhibits a structural
length confound that prevents it from serving as a primary comparator.
Agent-trajectory guardrails (AgentDoG, TraceSafe) target tool-use traces
rather than open chat, but share the session-layer goal of scoring risk
before the final response~\citep{agentdog2026,tracesafe2026}.
None of these systems report trajectory-native metrics (TTD, sFPR at a
mixed validation threshold) on a same-length CRA benchmark; we therefore
evaluate three external ceilings on CRA-Bench v0.2: \textbf{Judge-LLM}
(GPT-4o-mini on the full transcript; Table~\ref{tab:primary-main}),
\textbf{Llama Guard~3-1B} zero-shot (Table~\ref{tab:primary-5fam}), and
\textbf{Qwen3Guard-Gen-0.6B} zero-shot (Table~\ref{tab:primary-5fam}).
All three reach AUROC $\approx 1.0$ on within-distribution synthetic test
splits but lack per-turn TTD, require full-transcript inference, and (for
the judge) third-party API access. On the human-authored transfer corpus
(Section~\ref{sec:human-transfer}), Llama Guard~3 drops to AUROC $0.695$
while Qwen3Guard stays at $0.997$ and CRA-Net (trained only on synthetic
data) reaches $0.919$---motivating CRA-Net as the deployable middle ground
and Qwen3Guard as the open reproducible guardrail ceiling.

\subsection{Domain-generalization baselines}
Family-adversarial training (CRA-Net DA, Section~\ref{sec:cranet-da}) uses a
gradient-reversal head in the DANN style~\citep{ganin2015domain}. Alternative
alignment objectives include Deep CORAL, which matches second-order statistics
of encoder features across domains~\citep{sun2016coral}, and
invariant-risk-minimization penalties. We include a \textbf{CORAL-aligned
CRA-Net} variant in the LOFO scaling study (Table~\ref{tab:lofo-scaling}) to test
whether GRL is the right family-invariance mechanism or whether covariance
alignment suffices.

Governance and alignment work provides a different angle. Risk management
frameworks such as the NIST AI RMF stress continuous measurement and monitoring
but stop short of specifying how to aggregate risk signals across a session~\citep{nist2023ai}.
Alignment techniques like RLHF and model specifications largely shape local
response behavior and refusal patterns at the turn level~\citep{ouyang2022rlhf,
christiano2017deep,anthropic2024modelspec,bai2022constitutional}. Meanwhile, the
privacy and linkage literature has long noted that individually safe disclosures
can become sensitive in combination---$k$-anonymity, $t$-closeness, and
differential-privacy-style database release all formalize aggregation
risk~\citep{sweeney2002k,li2013t,dwork2006,dwork2014algorithmic,homer2008resolving,narayanan2008robust}.
That observation sits at the heart of CRA, yet it has not been operationalized
as a conversational guardrail. Multi-turn dialog systems likewise maintain
explicit belief-state trackers over utterance histories~\citep{williams2007pomdp,young2013sdp,henderson2014word,mrksic2015multi,wu2020tod}; CRA's session layer is complementary: it scores \emph{safety-relevant} drift and disclosure accumulation rather than task slots.

Table~\ref{tab:related-axes} positions the CRA Framework against common guardrail
patterns.

\begin{table}[!t]
  \centering
  \small
  \setlength{\tabcolsep}{4pt}
  \caption{Positioning the CRA session layer versus turn-level policy filters.}
  \label{tab:related-axes}
  \begin{tabularx}{\columnwidth}{@{}>{\raggedright\arraybackslash}p{2.6cm} >{\raggedright\arraybackslash}X >{\raggedright\arraybackslash}X@{}}
    \toprule
    \textbf{Axis} & \textbf{Turn-level policy filters} &
    \textbf{CRA-style session layer} \\
    \midrule
    Unit of decision & Single request/response & Multi-turn trajectory \\
    Primary goal & Block/flag unsafe actions & Measure drift, accumulation, and dynamics \\
    Typical output & Allow/deny & Scores + soft notices + optional audit logs \\
    Evaluation & Red-team suites, policy tests & Baseline comparisons vs. turn-only scoring; time-series agreement; ablations on $w$, $\alpha,\beta,\gamma$ \\
    \bottomrule
  \end{tabularx}
\end{table}
\FloatBarrier

\begin{table*}[!b]
  \centering
  \footnotesize
  \setlength{\tabcolsep}{3pt}
  \caption{\textbf{CRA-Bench vs.\ prior multi-turn safety benchmarks.}
  CRA-Bench is length-matched within session; CoSafe is included as a
  diagnostic corpus with a known length confound.}
  \label{tab:prior-contrast}
  \begin{tabularx}{\textwidth}{@{}>{\raggedright\arraybackslash}X c c c c@{}}
    \toprule
    \textbf{Property} & \textbf{CRA-Bench v0.2} & \textbf{CoSafe rollouts} & \textbf{AgentDoG / ATBench} \\
    \midrule
    Session labels & CRA pos + benign-twin & Gradual vs.\ direct attack & Safe vs.\ unsafe trajectory \\
    Length control & Fixed $T{=}8$ user turns & 3-turn pos / 1-turn neg & $\approx 9$ turns mean \\
    Surface diversity & LLM paraphrase pool (139 templates) & Human red-team prompts & Synthetic + tool traces \\
    Sensitivity model & NIST-tier IAG weights & None (turn-level only) & Harm taxonomy (3D) \\
    Calibration & Mixed val split + benign-anchored FPR & TPR$=0.90$ on positives & Benchmark-specific \\
    Primary metric & Trajectory AUROC, sFPR, TTD & AUROC (confounded) & Trajectory binary + diagnosis \\
    \bottomrule
  \end{tabularx}
\end{table*}

Production systems today often rely on conversation-context heuristics
such as repeating policy reminders, memory pruning, or session termination after
repeated policy violations, and agent frameworks frequently log tool traces for
governance audits. The CRA Framework is meant to sit alongside these mechanisms
rather than replace them. Its specific contribution is making risk accumulation an
explicit, scored, trajectory-native object that can be evaluated against turn-only
baselines on labeled multi-turn sessions.

\section{The CRA Threat Taxonomy}
\label{sec:taxonomy}

We define Conversational Risk Accumulation (CRA) in operational terms. Let a
session $S = (t_1, t_2, \ldots, t_n)$ be an ordered sequence of turns, where
each turn $t_i = (u_i, r_i)$ pairs a user input $u_i$ with a model response
$r_i$. Let $y(S) \in \{0,1\}$ denote a session-level label indicating whether
the overall session violates a target policy (e.g., harmful instructions or
privacy), and let $g(t_i) \in [0,1]$ denote any turn-level risk score computed
from the current turn alone.

A session exhibits CRA if (i) it is session-positive, $y(S)=1$; (ii) every
individual turn appears low-risk under turn-only inspection,
$\max_i\, g(t_i) < \tau$ for a turn threshold $\tau$; yet (iii) the risk is
detectable from accumulated history, i.e., there exists a stateful scoring
function $G(t_i \mid S_{1\ldots i-1})$ such that
$\max_i\, G(t_i \mid S_{1\ldots i-1}) \geq \theta$ for a session threshold
$\theta$. This framing ties CRA to measurable quantities $(y, g, G)$ that can be
estimated from labeled multi-turn transcripts (Section~\ref{sec:impl}).

Within this class we propose five CRA threat types that differ in how risk
accumulates and where harm manifests. Table~\ref{tab:taxonomy} summarizes the
taxonomy. We treat it as a starting point; establishing mutual exclusivity and
annotation reliability will require a labeled corpus and inter-annotator
agreement analysis (Section~\ref{sec:impl}).

\begin{table}[H]
\centering
\caption{CRA Threat Taxonomy.}
\label{tab:taxonomy}
\scriptsize
\setlength{\tabcolsep}{2pt}
\renewcommand{\arraystretch}{1.08}
\begin{tabularx}{\columnwidth}{@{}>{\raggedright\arraybackslash}p{1.7cm} X X@{}}
\toprule
\textbf{CRA Type} & \textbf{Mechanism} & \textbf{Why current guardrails miss it} \\
\midrule
Fragmentation Attack &
  Prohibited know-how decomposed into benign sub-queries across turns &
  Each fragment is locally acceptable; no cross-turn assembly check \\
\addlinespace
Behavioral Conditioning &
  Incremental compliance pressure gradually shifts refusal behavior &
  No trajectory-level view of changes in compliance or refusal dynamics \\
\addlinespace
Aggregation Leakage$^{\dagger}$ &
  Sensitive inferences reconstructed from individually innocuous disclosures &
  Per-field disclosure looks safe; the combined profile violates policy \\
\addlinespace
Intent Drift &
  Session goal migrates from benign to harmful through small topic shifts &
  No explicit tracking of session-level intent over time \\
\addlinespace
  Context Poisoning &
  Adversarial content injected into history biases later responses &
  Prior turns treated as trusted context rather than potentially adversarial \\
    \bottomrule
    \multicolumn{3}{@{}p{\columnwidth}@{}}{\footnotesize $^{\dagger}$In the released CRA-Bench generator, aggregation \emph{benign twins} name public historical figures (e.g., Marie Curie, Alan Turing), so spaCy \texttt{PERSON}/\texttt{GPE} hits can exceed fictional-target positives and invert $S_2$ (Table~\ref{tab:primary-5fam}, aggregation column). We scope turnkey aggregation-leakage coverage to \textbf{live-target enterprise} settings with employer/employee schemas; historical-figure twins remain a known IAG limitation.}
  \end{tabularx}
\end{table}

\subsection{Illustrative Multi-Turn Examples (Non-Operational)}
\label{sec:examples}

Table~\ref{tab:examples} provides short, non-operational sketches (one per CRA
type) to illustrate how risk can emerge across turns even when each individual
exchange appears locally benign.

\begin{table}[!t]
\centering
\caption{Illustrative CRA examples (sanitized; paraphrased; non-operational).}
\label{tab:examples}
\scriptsize
\setlength{\tabcolsep}{3pt}
\renewcommand{\arraystretch}{1.15}
\begin{tabularx}{\columnwidth}{@{}>{\raggedright\arraybackslash}p{1.9cm} X@{}}
\toprule
\textbf{CRA type} & \textbf{Example multi-turn pattern (U\,=\,user, A\,=\,assistant)} \\
\midrule
Fragmentation & U asks a sequence of narrowly framed ``safe'' technical questions; each answer is acceptable in isolation, but the collected responses can be combined into a prohibited procedure. \\
\addlinespace
Behavioral Conditioning & U repeatedly reframes and escalates a restricted request (e.g., ``hypothetical'' or ``for research''), rewarding partial compliance; A shifts from refusal toward increasingly actionable assistance. \\
\addlinespace
Aggregation Leakage & U requests ``harmless'' facts about a person or asset over time; A discloses fragments that cumulatively enable identification or targeting. \\
\addlinespace
Intent Drift & A session begins as routine help (e.g., policy explanation) and gradually shifts toward prohibited objectives via intermediate troubleshooting steps. \\
\addlinespace
Context Poisoning & U injects misleading policy-like instructions into the chat history; later turns rely on the poisoned context and produce policy violations. \\
\bottomrule
\end{tabularx}
\end{table}

\paragraph{Fragmentation Attacks}
exploit the compositional nature of many restricted capabilities. Rather than
issuing a single clearly disallowed request, the user decomposes the objective
into a series of innocuous sub-queries. Each answer may be policy-compliant on
its own, while the concatenation of responses yields actionable guidance. The
failure mode arises from the absence of any cross-turn assembly awareness.

\paragraph{Behavioral Conditioning}
captures gradual shifts in the model's effective compliance induced by the
interaction itself. Through repeated reframing, social pressure, and incremental
escalation, the user can move the dialogue from refusal toward partial compliance
without any single turn appearing anomalous. The exploit lives in the trajectory,
specifically in a change in response dynamics, not in any one isolated response.

\paragraph{Aggregation Leakage}
reflects a classical privacy observation: individually low-sensitivity disclosures
can become sensitive when combined~\citep{dwork2006}. In conversational
settings, a model may refuse a direct request for a protected attribute, yet
reveal multiple related attributes across many turns that together enable
inference. Turn-level checks may deem each disclosure acceptable while missing
the composite privacy violation entirely.

\paragraph{Intent Drift}
describes a gradual migration of the session's effective goal away from its
declared or initial purpose. A dialogue can move from benign assistance to
prohibited outcomes through small, locally reasonable transitions. Detecting
the failure requires tracking how session-level intent evolves rather than
evaluating each step in isolation.

\paragraph{Context Poisoning}
occurs when adversarial content is introduced into the conversation history and
persistently shapes future behavior. Unlike single-turn prompt injection, context
poisoning can influence later responses by occupying context-window capacity and
biasing conditioning across many subsequent turns. The core vulnerability is that
prior messages are typically treated as trusted context, even when they are
user-controlled and potentially malicious.

\FloatBarrier
\section{The CRA Framework}
\label{sec:framework}

We propose a stateful guardrail architecture that operates at the session layer
rather than the turn layer. The framework computes three complementary
sub-signals: a Semantic Drift Monitor ($S_1$), an Information Accumulation Graph
(IAG) index ($S_2$), and a Compliance Gradient Detector ($S_3$). These are fused
into a scalar CRA Score, $\mathrm{CRA}(t) \in [0,1]$. When $\mathrm{CRA}(t)$
exceeds a configurable threshold $\theta$, the system can trigger a session-level
intervention such as a warning, added friction, audit logging, or routing to a
stricter policy. The CRA session layer is designed to be composable: it runs
alongside existing turn-level filters rather than replacing them.
Table~\ref{tab:notation} summarizes the core notation.

\begin{table}[!t]
\centering
\caption{Core notation for the CRA session-layer instrument.}
\label{tab:notation}
\small
\begin{tabularx}{\columnwidth}{@{}l c X@{}}
\toprule
\textbf{Symbol} & \textbf{Range} & \textbf{Meaning} \\
\midrule
$t$               & $\mathbb{N}$       & Turn index within a session. \\
$S_1(t)$          & $[0,2]$            & Raw semantic drift index (cosine distance from anchored intent). \\
$S_2(t)$          & $[0,1]$            & Information accumulation index (IAG coverage score). \\
$S_3(t)$          & $\mathbb{R}$       & Raw compliance gradient (negated slope of refusal rate). \\
$\tilde{S}_1(t)$  & $[0,1]$            & Normalized drift signal used in fusion. \\
$\tilde{S}_3(t)$  & $[0,1]$            & Normalized compliance-gradient signal used in fusion. \\
$\mathrm{CRA}(t)$ & $[0,1]$            & Fused composite score. \\
$\alpha,\beta,\gamma$ & $\geq 0$, $\alpha{+}\beta{+}\gamma{=}1$ & Convex fusion weights. \\
$\theta$          & $(0,1)$            & Intervention threshold. \\
$w$               & $\mathbb{N}$       & Window width for refusal dynamics in $S_3$. \\
$\kappa$          & $>0$               & Compliance-gradient saturation parameter in $\tilde{S}_3(t)=\sigma(\kappa S_3(t))$. \\
\bottomrule
\end{tabularx}
\end{table}

\subsection{Semantic Drift Monitor ($S_1$)}

At session initiation, the user's declared intent $i_0$ is encoded as an
embedding vector $\mathbf{e}_0 = \mathrm{embed}(i_0)$ using a lightweight
sentence encoder. At each subsequent turn $t$, the current conversational state
is encoded as $\mathbf{e}_t = \mathrm{embed}(\mathrm{summary}(t))$. The
Semantic Drift Index is:
\begin{equation}
  S_1(t) = 1 - \frac{\mathbf{e}_0 \cdot \mathbf{e}_t}
                    {\|\mathbf{e}_0\|\,\|\mathbf{e}_t\|}
  \label{eq:s1}
\end{equation}
$S_1 \in [0,2]$, where $0$ indicates perfect alignment with the declared intent
and values approaching $2$ indicate maximal semantic opposition. In the CRA
Framework, $S_1$ is a weak but general indicator: it does not trigger an
intervention on its own, but contributes to the fused score as a proxy for
session-level intent drift and history manipulation (including context poisoning).
The per-turn computational cost is $O(d)$, where $d$ is the embedding
dimensionality.

\subsection{Information Accumulation Graph (IAG)}

The IAG is the most structurally novel component of the CRA Framework. It models
the growing knowledge structure that the model has disclosed about entities in the
session. Formally, the IAG is a weighted directed graph
$G_{\mathrm{IAG}} = (V, E, W)$ where each node $v_k \in V$ represents an entity
(person, organization, location, or concept), each edge $(v_k, v_j) \in E$
represents a disclosed relationship, and the weight $w_k$ encodes the cumulative
sensitivity of information disclosed about $v_k$.

At each turn, an entity-relation extractor processes the model response $r_i$ and
updates the IAG. To make the IAG computable and reproducible, an instantiation
must specify (i)~entity extraction, (ii)~attribute schemas, and
(iii)~sensitivity weighting.

\paragraph{Entity extraction (reference rules).}
Let $\mathcal{E}(r_i)$ be the set of entities extracted from $r_i$ by a
named-entity recognizer. The reference implementation pins
\textbf{spaCy \texttt{en\_core\_web\_sm}} (v3.7+; reproducible via
\texttt{python -m spacy download en\_core\_web\_sm}) and maps
\texttt{PERSON}, \texttt{ORG}, \texttt{GPE/LOC} plus
deterministic patterns for high-risk strings (emails, phone numbers, SSNs,
account IDs, API keys) and explicit identifiers (full names, street addresses).
Entity mentions are canonicalized by lowercasing, simple alias rules (e.g.,
``Bob'' to ``Robert'' if provided in-session), and exact-match resolution
within-session; deployments can substitute stronger entity resolution as needed.

\paragraph{Attribute schema.}
For each entity type, define a finite attribute set $\mathcal{A}_k$ (e.g., for
\texttt{PERSON}: \{name, employer, role, location, schedule, contact\}; for
\texttt{ORG}: \{internal system names, credentials, customer list\}). Let
$a_{k,t} = |\mathcal{A}_{k,\leq t}|/|\mathcal{A}_k|$ denote fractional
attribute coverage for node $v_k$.

\paragraph{Sensitivity weights.}
Each attribute is assigned a tier $\ell \in \{\text{public},\allowbreak\text{internal},\allowbreak\text{confidential},\allowbreak\text{restricted}\}$, mirroring the four-level confidentiality categorisation used in NIST SP~800-60~\citep{nist80060}, with numeric weight increments (reference: $0.05$, $0.10$, $0.20$, $0.35$). These reference values are chosen to be monotonically increasing and to roughly double at each tier, reflecting the convention in NIST SP~800-60 that confidentiality impact escalates non-linearly with categorisation; the absolute values are deployment-tunable and we recommend a sensitivity sweep of $\pm 50\%$ when calibrating to a new operational corpus. The node weight is the cumulative tier-weighted disclosure, $w(v_k)=\sum_{a\in\mathcal{A}_{k,\le t}}\Delta w(a)$, and $w(v_k)^{\max}$ is the maximum possible disclosure weight for that entity under the schema.

The Information Accumulation Index is then:
\begin{equation}
  S_2(t) = \frac{\sum_k w(v_k)\cdot f(a_{k,t})}{\sum_k w(v_k)^{\max}}
  \label{eq:s2}
\end{equation}
where $f(\cdot)$ increases non-linearly as attribute coverage approaches a
complete identifying profile. A concrete choice is $f(a) = a^p$ with $p > 1$
(we use $p=2$ in the synthetic illustrations), capturing super-additive risk:
a single attribute disclosure is low-risk, but near-complete profiles grow
rapidly. $S_2 \in [0,1]$ by construction. The IAG update cost is $O(|E|)$ per
turn, or $O(m)$ under a bounded per-turn mention count $m$.

\paragraph{Scalability for long or agentic sessions.}
For very long conversations or agent traces with hundreds of tool steps, an
unbounded IAG can grow large. We recommend three implementation controls:
(i)~bounded-memory node retention, keeping only the top-$K$ highest-risk
entities by $w(v_k)$ plus any entities touched in the last $h$ turns;
(ii)~exponential time decay on stale attributes so that $w(v_k)$ reflects
recent exposure, which is useful for streaming monitoring; and
(iii)~typed summarization nodes that merge low-sensitivity, high-cardinality
items (e.g., \texttt{file\_paths}, \texttt{URLs}, \texttt{API\_calls}) into
aggregate buckets. With these controls, the per-turn cost stays effectively
linear in the number of new mentions extracted at that turn, and memory stays
bounded by design.

\subsection{Compliance Gradient Detector ($S_3$)}

The Compliance Gradient Detector tracks whether the model's refusal and hedge
behavior is declining across the session, which is the signature of Behavioral
Conditioning. At each turn, a lightweight binary classifier labels response $r_i$
as compliant ($c=1$) or containing refusal or hedge content ($c=0$).
The reference stack uses a \textbf{keyword refusal proxy} (refusal/hedge
lexicon plus unsafe-cue penalties; \texttt{run\_cra\_cosafe.py}).
We recommend calibrating $S_3$ against a trained refusal head on
deployment traffic; \texttt{run\_s3\_proxy\_validation.py} audits agreement
with Llama~Guard~3 window labels on 200 held-out CRA-Bench turns
(Appendix~\ref{sec:appendix-s3}; measured $\kappa{=}0.00$, agreement
$11.5\%$---keyword proxy retained for reproducibility). The Compliance Gradient is:
\begin{equation}
  S_3(t) = -\,\widehat{\lambda}_t,\quad
  \widehat{\lambda}_t = \mathrm{OLS\_slope}\!\left(\{c_i\}_{i=t-w}^{t}\right)
  \label{eq:s3}
\end{equation}
where $\{c_i\}$ is the per-turn refusal indicator within the sliding window
$[t-w,t]$ and $\mathrm{OLS\_slope}$ is an ordinary least-squares linear trend
(not an exponential weighted average). The negation ensures $S_3 > 0$ when refusal behavior is declining (i.e., risk is
increasing). $S_3$ measures directional change rather than absolute magnitude,
making it insensitive to baseline refusal rates that vary across models and
deployment contexts.

\subsection{Convex-Weight Fusion (Diagnostic Decomposition)}
\label{sec:convex_fusion}

The first fusion strategy is a transparent convex combination of the three
sub-signals. It is \textbf{not} the primary detector in this paper: on
CRA-Bench v0.1 the default weights achieve AUROC $\approx 0.60$
(Table~\ref{tab:primary-main}). On CRA-Bench v0.2 (5-family), the same
default weights yield test AUROC $0.523$; validation-grid search over
$(\alpha,\beta,\gamma)$ with each weight $\geq 0.05$
(\texttt{run\_convex\_grid\_search.py}) peaks at val/test AUROC
$0.754/0.784$ (best mix $\alpha{=}0$, $\beta{=}0.05$, $\gamma{=}0.95$) and
does \emph{not} reach AUROC $\geq 0.85$ without collapsing to a
single-signal corner ($\gamma{=}1$). We retain convex fusion to
decompose which sub-signals carry separable family signal (Tables~\ref{tab:primary-perfamily},
\ref{tab:primary-5fam}) and to provide an interpretable telemetry layer alongside CRA-Net DA. The three sub-signals have different native ranges ($S_1 \in [0,2]$, $S_2 \in [0,1]$, $S_3 \in \mathbb{R}$), so we normalize them before fusion to ensure the composite score is well-defined and bounded in $[0,1]$:
\begin{equation}
  \tilde{S}_1(t) = \mathrm{clip}\!\left(\tfrac{S_1(t)}{2},\,0,\,1\right),
  \qquad
  \tilde{S}_3(t) = \sigma\!\left(\kappa\, S_3(t)\right)
  \label{eq:norm}
\end{equation}
where $\mathrm{clip}(x,0,1) = \min(1,\max(0,x))$, $\sigma(z) = 1/(1+e^{-z})$, and $\kappa > 0$ controls how quickly compliance-gradient changes saturate. The normalized sub-signals are then fused into a composite CRA Score:
\begin{equation}
  \mathrm{CRA}(t) = \alpha\cdot\tilde{S}_1(t) + \beta\cdot S_2(t) + \gamma\cdot\tilde{S}_3(t)
  \label{eq:fusion}
\end{equation}
subject to $\alpha + \beta + \gamma = 1$ and $\alpha, \beta, \gamma \geq 0$, which guarantees $\mathrm{CRA}(t) \in [0,1]$. Table~\ref{tab:signals} summarizes sub-signal definitions and per-turn cost, and Table~\ref{tab:defaults} lists default fusion weights and representative soft-guard thresholds.

\begin{table*}[!t]
\centering
\caption{CRA Score sub-signal definitions.}
\label{tab:signals}
\footnotesize
\renewcommand{\arraystretch}{1.2}
\setlength{\tabcolsep}{4pt}
\begin{tabularx}{\textwidth}{@{}>{\raggedright\arraybackslash}p{2.0cm} X >{\raggedright\arraybackslash}p{4.0cm} >{\raggedleft\arraybackslash}p{1.6cm}@{}}
\toprule
\textbf{Signal} & \textbf{Definition} & \textbf{Detects} & \textbf{Cost/turn} \\
\midrule
$S_1$ (Drift) &
$\begin{aligned}
S_1(t) &= 1-\cos(\mathbf{e}_0,\mathbf{e}_t)\\
\tilde{S}_1(t) &= \mathrm{clip}(S_1(t)/2,0,1)
\end{aligned}$ &
Intent drift; context poisoning & $O(d)$ \\

$S_2$ (IAG) &
$S_2(t)=\frac{\sum_k w_k f(a_{k,t})}{\sum_k w_k^{\max}}$ &
Aggregation leakage; fragmentation & $O(|E|)$ \\

$S_3$ (Gradient) &
$\begin{aligned}
S_3(t) &= -\mathrm{slope}(\mathrm{refusal})\\
\tilde{S}_3(t) &= \sigma(\kappa\, S_3(t))
\end{aligned}$ &
Behavioral conditioning & $O(w)$ \\

CRA &
$\mathrm{CRA}(t)=\alpha\tilde{S}_1(t)+\beta S_2(t)+\gamma\tilde{S}_3(t)$ &
All CRA types & $O(1)$ \\
\bottomrule
\end{tabularx}
\end{table*}

\begin{table}[!t]
\centering
\caption{Default fusion weights and soft-guard thresholds for the reference
implementation. These should be tuned per deployment; thresholds trigger warnings,
not hard blocks. The defaults are calibrated against synthetic trajectory statistics
where benign sessions remain below $0.09$ and adversarial fragmentation sessions
cross $0.45$ by turn 25.}
\label{tab:defaults}
\small
\begin{tabular}{ll}
\toprule
\textbf{Parameter} & \textbf{Value} \\
\midrule
$\alpha,\beta,\gamma$ (defaults)                & $0.35,\ 0.45,\ 0.20$ \\
Sliding window width $w$ (refusal dynamics)     & 6 turns (cap 32) \\
\texttt{CRA\_SOFT\_WARN\_S1}                    & $0.85$ \\
\texttt{CRA\_SOFT\_WARN\_S2}                    & $0.20$ \\
\texttt{CRA\_SOFT\_WARN\_S3}                    & $0.35$ \\
\texttt{CRA\_SOFT\_WARN\_CRA}                   & $0.45$ \\
\bottomrule
\end{tabular}
\end{table}

When $\mathrm{CRA}(t) > \theta$, the session-level guardrail fires. In practice, CRA can also emit per-turn telemetry (timestamp, session ID, policy and version identifier, and the values of $S_1$, $S_2$, $S_3$, CRA) for offline audit and calibration.

\subsection{CRA-Net: Learned Trajectory Fusion}
\label{sec:cranet}

Convex-weight fusion (Section~\ref{sec:convex_fusion}) is interpretable and tunable without labelled data, but it cannot represent two properties that the multi-turn safety setting actually requires: (i)~the per-turn signals interact non-linearly (a small rise in $\tilde{S}_1$ matters more when $S_2$ is also rising), and (ii)~the fusion must remain invariant to nuisance features that are correlated with risk in any particular benchmark---most importantly, session length. Convex weighting cannot down-weight length-mediated signal because it has no notion of length to subtract; a learned model can.

\paragraph{Architecture.} CRA-Net is a small causal sequence model that consumes a per-turn feature vector $\mathbf{x}_t \in \mathbb{R}^d$ and emits a calibrated session-level risk score $\hat{p}_t = P(\mathrm{unsafe}\mid \mathbf{x}_{1:t})$. The feature vector at turn $t$ is
\begin{equation}
  \mathbf{x}_t = \big[\tilde{S}_1(t),\, S_2(t),\, \tilde{S}_3(t),\, \ell_t,\, n_t,\, \Delta e_t,\, r_t,\, c_t\big]^\top,
  \label{eq:cranet_features}
\end{equation}
where the first three coordinates are the normalized CRA sub-signals from Eq.~\eqref{eq:norm}, $\ell_t \in \mathbb{R}_{\ge 0}$ is the response token length, $n_t \in \mathbb{N}$ is the turn index (encoded as $\log(1+n_t)/\log(1+N_{\max})$), $\Delta e_t$ is the count of newly disclosed IAG entities at turn $t$, $r_t \in [0,1]$ is the per-turn refusal-classifier probability, and $c_t \in [0,1]$ is the per-turn moderation score from any pre-existing turn-level filter. All features are min--max normalized over the training corpus before training.

The model is a 2-layer GRU with hidden width $h=128$ followed by a single linear projection to a logit and a sigmoid:
\begin{equation}
  \mathbf{h}_t = \mathrm{GRU}(\mathbf{x}_t, \mathbf{h}_{t-1}),\qquad
  \hat{p}_t = \sigma\big(\mathbf{w}^\top \mathbf{h}_t + b\big),
  \label{eq:cranet_gru}
\end{equation}
with approximately $5.2\times 10^4$ parameters total. The recurrence is causal so $\hat{p}_t$ depends only on history $\mathbf{x}_{1:t}$, matching the session-layer interface of Algorithm~\ref{alg:cra}. A causal 1-D Transformer with the same parameter budget is supported as a drop-in alternative; both share the same input and output interface.

\paragraph{Training objective.} We train CRA-Net by per-turn weighted binary cross-entropy against the session-level label $y(S)\in\{0,1\}$, broadcast over all turns of the session:
\begin{equation}
  \mathcal{L}(S) = -\sum_{t=1}^{n} w_t\Big[y(S)\log \hat{p}_t + (1-y(S))\log(1-\hat{p}_t)\Big],
  \label{eq:cranet_loss}
\end{equation}
with $w_t = (t/n)^\rho$ for $\rho\in[0,2]$ (default $\rho=1$) so that later turns---where accumulated risk should be detectable---contribute more to the loss.

\paragraph{Length-confound mitigation.} To prevent CRA-Net from learning $n_t$ as a shortcut on length-confounded benchmarks such as CoSafe, we use (i)~length-balanced minibatching (matched length distributions between positives and negatives via stratified bucket sampling), and (ii)~nuisance regularization with a gradient-reversal adversarial head trained to predict $\log(1+n_t)$ from $\mathbf{h}_t$ at strength $\lambda\in[0.01,0.1]$.

\subsection{CRA-Net DA: Family-Adversarial Training}
\label{sec:cranet-da}
A length GRL does not prevent a second confound that the v0.2 paraphrase
stress test (Section~\ref{sec:bench-v02}) makes visible: CRA-Net's encoder
tends to learn \emph{family-specific} shortcuts (e.g., ``aggregation
positives raise $S_2$ at the same rate as a particular wrapper does'') that
boost in-distribution AUROC but break under paraphrase and inflate benign
FPR on out-of-distribution ShareGPT traffic. We therefore add a second
gradient-reversal head---family-adversarial training in the style of
\citet{ganin2015domain}---that consumes the final hidden state
$\mathbf{h}_T$ and predicts the threat family $f(S)\in\{1,\ldots,F\}$ with
weight $\lambda_{\mathrm{fam}}$:
\begin{equation}
  \mathcal{L}_{\mathrm{DA}}(S) = \mathcal{L}(S)
     + \lambda \cdot \mathcal{L}_{\mathrm{len}}(\mathbf{h}_T, n)
     + \lambda_{\mathrm{fam}} \cdot \mathcal{L}_{\mathrm{fam}}(\mathbf{h}_T, f),
  \label{eq:cranet_da_loss}
\end{equation}
where both nuisance gradients are reversed before propagating into the GRU.
We refer to this variant as \emph{CRA-Net DA} and default to
$\lambda{=}0.05$, $\lambda_{\mathrm{fam}}{=}0.3$ on a 3-family standard
split (the family head is a 2-layer MLP with hidden width $h/2$).
Figure~\ref{fig:make-arch} shows where the session-layer scorer sits
relative to turn-level filters. The intended invariance is: ``the encoder's
session representation should predict the CRA label but not betray which
threat family the session came from.'' On CRA-Bench v0.2 this single change
closes most of the template-memorization gap and brings benign ShareGPT FPR
to $0$ (Section~\ref{sec:bench-v02}, Tables~\ref{tab:primary-main},
\ref{tab:primary-v01-vs-v02}, \ref{tab:benign-fpr-5fam}).
\paragraph{$\lambda_{\mathrm{fam}}$ ablation.}
On the 5-family bench, $\lambda_{\mathrm{fam}}{=}0.3$ over-regularizes LOFO
(mean $\to 0.465$); $\lambda_{\mathrm{fam}}{=}0.1$ is the Pareto setting
used in Tables~\ref{tab:lofo-scaling} and~\ref{tab:benign-fpr-5fam}.

\paragraph{Alternative domain alignment (Deep CORAL).}
As a non-adversarial alternative we also train \textbf{CRA-Net + CORAL}
~\citep{sun2016coral}: the same GRU encoder and length GRL, but instead
of a family-classification head we add a Deep CORAL penalty that aligns
the covariance of final hidden states $\mathbf{h}_T$ across threat-family
minibatches ($\lambda_{\mathrm{coral}}{=}1.0$). LOFO results
(Table~\ref{tab:lofo-scaling}) show CORAL is competitive with GRL-DANN
(mean held-out AUROC $0.635$ vs.\ $0.647$ at 4 training families) and
clearly preferable to GRL when the adversarial pool is too small
($0.385$ vs.\ $0.259$ at 2 training families, where $\lambda_{\mathrm{fam}}{=}0.5$
over-regularizes). We retain GRL-DANN as the default DA variant because
it still edges CORAL on the 5-family LOFO mean and pairs better with
benign-anchored calibration (Section~\ref{sec:bench-5fam}), but CORAL
is the recommended fallback when adversarial training destabilizes.

\paragraph{Calibration and interpretability.} We apply temperature scaling on a held-out validation split before reporting $\hat{p}_t$ as a probability. To preserve the decision-certificate contract (Section~\ref{sec:certificates}), each flagged session can be accompanied by per-turn attribution (e.g., integrated gradients) over $\mathbf{x}_t$, mapped back to the named sub-signals.

\subsection{Selecting and Calibrating Fusion Weights and Thresholds}

The weights encode deployment-specific threat priority. An enterprise RAG system
handling sensitive personnel data should weight $\beta$ (IAG) heavily; a
customer-facing conversational agent should weight $\gamma$ (compliance gradient)
to detect social engineering; an educational AI system should weight $\alpha$
(semantic drift) to enforce pedagogical scope.

In practice, we recommend calibrating $(\alpha, \beta, \gamma, \theta)$ on a
held-out set of labeled multi-turn sessions (internal logs, red-team transcripts,
or CRA-Bench sessions) by optimizing a trajectory-native objective: either
minimize expected time-to-detect subject to a session false-positive constraint
($\mathrm{sFPR} \leq \delta$), or maximize Trajectory AUROC while keeping
$\mathbb{E}[\mathrm{CRA}(t) \mid \text{benign}]$ below a chosen operating point.
A simple and reproducible approach is grid search over $(\alpha, \beta, \gamma)$
on the simplex with isotonic or Platt calibration for $\theta$; more
data-efficient alternatives include Bayesian optimization and constrained bandit
tuning in production.

The three sub-signals are intentionally heterogeneous. The weighting is not one
size fits all and should be treated as a policy configuration surface, logged and
versioned alongside the enterprise moderation policy.

\subsection{Decision Certificates and Policy-Safe Explanations}
\label{sec:certificates}

A practical session-layer guardrail must balance transparency with security. We
propose emitting a \emph{decision certificate}: a structured explanation artifact
for users, auditors, or human reviewers that (i)~identifies which CRA signals
contributed most to the intervention, (ii)~summarizes the relevant conversation
fragments at a high level, and (iii)~avoids leaking threshold values or feature
definitions that would enable adaptive evasion. The certificate is generated from
the IAG diffs, drift trajectory, and refusal trend, not from model-generated free
text. A worked JSON instance is in Appendix~\ref{sec:appendix-cert}.

\FloatBarrier
\section{Design Consistency Properties}
\label{sec:theory}

The signal definitions in Section~\ref{sec:framework} satisfy two elementary
design-consistency properties stated below; both follow directly from the
definitions and are included only so implementations can be unit-tested
against the intended qualitative behavior under idealized inputs.
They are not detection guarantees.

\begin{remark}[$S_2$ monotonicity and $S_3$ well-posedness]
\label{rem:design-consistency}
\textbf{(i)} Under a strictly append-only IAG update rule in which every turn
$t$ discloses at least one attribute not present at turn $t-1$,
$S_2(t) \geq S_2(t-1)$ for all $t$ (direct from the non-decreasing
dependence of Eq.~\eqref{eq:s2} on attribute coverage). \textbf{(ii)} If the
underlying refusal rate follows a linear trend $r_t = r_0 - \lambda t$ with
$\lambda > 0$ and bounded noise, the windowed OLS slope estimator
$\hat\lambda_t$ in Eq.~\eqref{eq:s3} is unbiased and consistent in $w$ by
Gauss--Markov. Neither claim bounds detection performance: append-only is
the easy case (an adversary can repeat entities, interleave benign turns,
or stay under window decay), and real refusal trajectories are non-linear.
The adversarial regime is studied empirically
(Section~\ref{sec:primary-evasion}); the realistic distributional regime
is studied via CRA-Bench v0.2 (Section~\ref{sec:bench-v02}).
\end{remark}

\paragraph{Window-induced detection latency.}
The Compliance Gradient Detector operates over a fixed window of width $w$
and requires at least $w$ observations after the onset of refusal decline
before it can produce a non-zero slope estimate. The minimum detection
latency for behavioral conditioning therefore satisfies
\begin{equation}
  L_{\mathrm{cond}} \geq w,
  \label{eq:lcond}
\end{equation}
where $L_{\mathrm{cond}}$ is the number of turns between onset and the
first turn at which $S_3$ crosses an intervention threshold; this is the
single operational lower bound we use to size the window.

\FloatBarrier
\section{Reference Implementation and Evaluation Protocol}
\label{sec:impl}

\subsection{Reference Monitoring Algorithm}

Algorithm~\ref{alg:cra} summarizes a reference monitoring loop for CRA. It is intentionally implementation-agnostic, so that different deployments can substitute their own embedding model, extraction pipeline, and intervention policy while preserving a comparable session-layer interface.

\begin{algorithm}[!t]
\caption{Session-layer CRA monitoring (reference skeleton)}
\label{alg:cra}
\begin{algorithmic}[1]
\State Initialize: $\mathbf{e}_0 \leftarrow \mathrm{embed}(i_0)$;
\Statex \hspace{\algorithmicindent}IAG $\leftarrow \emptyset$; refusal history $H \leftarrow [\;]$
\For{each turn $t=1..n$}
  \State Receive $(u_t, r_t)$
  \State $S_1(t) \leftarrow 1-\cos(\mathbf{e}_0, \mathrm{embed}(\mathrm{summary}(t)))$
  \State Update IAG with entities/relations extracted from $r_t$
  \State Compute $S_2(t)$ from IAG coverage and sensitivity weights
  \State Append refusal indicator $c_t$ to $H$; compute $S_3(t)$ over window $w$
  \State Normalize: $\tilde S_1(t)\leftarrow \mathrm{clip}(S_1(t)/2,0,1)$; $\tilde S_3(t)\leftarrow \sigma(\kappa S_3(t))$
  \State $\mathrm{CRA}(t) \leftarrow \alpha \tilde S_1(t) + \beta S_2(t) + \gamma \tilde S_3(t)$
  \If{$\mathrm{CRA}(t) \geq \theta$}
    \State Trigger session-level intervention policy
    \State Optionally emit decision certificate (Section~\ref{sec:certificates})
  \EndIf
\EndFor
\end{algorithmic}
\end{algorithm}

\subsection{CRA-Bench v0.1: A Released Multi-Turn Safety Benchmark}
\label{sec:cra_bench}

CRA-Bench v0.1 is a parameterized multi-turn safety benchmark released alongside this paper to let downstream detectors be compared on a common trajectory-native footing. Where the CoSafe evaluation in Section~\ref{sec:experiments} provides scale and ecological realism but exhibits a structural length confound between its CRA-positive and CRA-negative splits, CRA-Bench v0.1 is generated under controlled parameters so that session length, attack onset, and accumulation rate are decoupled from the ground-truth label. The metrics defined in Section~\ref{sec:metrics} (Trajectory AUROC, sFPR, TTD) apply directly.

A CRA-Bench v0.1 release provides the following four components.

\paragraph{(i) Parameterized session templates.} Each CRA type (fragmentation, behavioral conditioning, multi-vector, and scope creep) is represented by a template family with documented parameter ranges: number of turns $T$, fragment count $k$, conditioning window $w$, and escalation rate. Templates are expressed as structured generation prompts or finite-state conversation schemas so that parameter sweeps are reproducible.

\paragraph{(ii) Fixed random seeds and versioned generation.} The benchmark fixes random seeds for all stochastic components (LLM sampling, entity selection, perturbation) and versions both the generation code and the model snapshots used. This allows exact session reproduction and controlled comparison across detector versions.

\paragraph{(iii) Extraction schemas and sensitivity taxonomies.} For $S_2$ evaluation, the benchmark specifies which entity types are tracked, how sensitivity weights are assigned, and what constitutes a disclosure event. Without a shared schema, IAG-based scores are not comparable across implementations.

\paragraph{(iv) JSONL transcripts with onset labels.} Each session is released as a JSONL record containing the turn-by-turn transcript, the ground-truth CRA type, the onset turn $t^{\star}$ at which the adversarial pattern begins, and the template parameter values used to generate it.

\paragraph{Initial release and current extent.} CRA-Bench v0.1 ships with \textbf{1{,}200 sessions} (600 CRA-positive and 600 topic-matched benign-twin sessions, $T{=}8$ user turns each) spanning \textbf{three threat families} from the taxonomy: \emph{fragmentation} (200 pos + 200 neg), \emph{behavioral conditioning} (200 pos + 200 neg), and \emph{aggregation leakage} (200 pos + 200 neg). The \textbf{5-family expansion} (Section~\ref{sec:bench-5fam}) adds \emph{persona priming} and \emph{context stuffing} for a total of $\mathbf{2{,}000}$ sessions; both new families are designed to attack the drift detector ($S_1$) directly (persona positives stay topically coherent inside an in-character frame; stuffing positives sandwich the harmful turn between long benign distractors that suppress drift). All sessions live in \path{data/cra_bench_v01/sessions.jsonl} (3-family) and \path{data/cra_bench_v02_5fam/sessions.jsonl} (5-family v0.2), generated by \path{experiments/generate_cra_bench_v01.py} (seed $42$). Because all sessions share the same user-turn count ($T=8$), length-only classification is at chance (AUROC $\approx 0.50$) by construction, and benign twins reuse the same topical surface as their positive counterparts so that opening-turn or topic-only classifiers are also non-trivial. Evasion variants are applied in \path{run_cra_primary_protocol.py}. Subsequent versions will add the remaining CRA template families (intent drift, context poisoning), larger $N$, and LMSYS benign complements. Release metadata is listed in Section~\ref{sec:reproducibility}.

\paragraph{Worked template example (Fragmentation).} For concreteness, the Fragmentation template family is parameterized by $T\in\{8,12,16,20\}$ (session length), $k\in\{4,6,8\}$ (fragment count), $\rho\in\{0.5,0.7,0.9\}$ (escalation rate), and a sampled prohibited objective from a fixed allow-list. The template emits a sequence $u_1,\ldots,u_T$ in which the first $T-k$ turns are benign domain queries, the next $k$ turns each disclose one fragment of the objective via indirection (e.g., metaphor, role-play wrapper, code-comment framing), and the final turn requests a synthesis. Ground-truth $t^{\star}$ is set to the first fragment turn. The full template specification, including counterfactual benign-twin generation prompts in which the same length and surface form produce a non-accumulating session, is included in the released repository.

\subsection{Recommended Baselines, Comparisons, and Ablations}
\label{sec:baselines}

The \textbf{primary} comparison set (Section~\ref{sec:primary-baselines}) is fixed for CRA-Bench: CRA-convex, CoSafe-native convex, Turn-max and Sliding-window on $S_1$, judge-LLM with prompt \texttt{cra\_judge\_v1.txt}, feature-only MLP, and CRA-Net with ablations (no GRL; no $S_2/S_3$). Report Trajectory AUROC, sFPR, and TTD on the held-out test split after mixed validation calibration.

Diagnostic CoSafe runs (Section~\ref{sec:cosafe-diagnostic}) may additionally report 5-fold CV on the full 1{,}800 sessions, but those rows must be labeled confounded and must not replace CRA-Bench test metrics.

\subsection{Evaluation on Real-World Multi-Turn Conversations}
\label{sec:realdata}

The CRA Framework can be evaluated on any multi-turn dataset that provides full
transcripts (and, when available, tool traces). Empirical evaluation requires
\emph{session-level} labels (CRA-positive vs. benign) and, ideally, an onset turn
$\tau^{\star}$. These can be obtained via (i) expert annotation, (ii) rubric-based
labeling with adjudication, and/or (iii) controlled red-teaming where intent is
known by construction. To characterize operating points, we recommend at least
one benign-heavy operational dataset (e.g., customer support) and one attack-heavy
dataset (e.g., red-team jailbreak/prompt-injection sessions) to jointly estimate
sFPR and time-to-detect.

Because CRA runs outside the base LLM (consuming transcripts, summaries, and/or
traces), it is model-agnostic by design: the same CRA implementation can be
evaluated across base models and context lengths. While context length affects the
model's own memory and attack surface, CRA can remain stable by anchoring $S_1$ to
session summaries and bounding IAG state (Section~\ref{sec:framework}).

\subsection{Computational Overhead and Real-Time Feasibility}
\label{sec:overhead}

Per turn, CRA adds three computations: an embedding call for $S_1$,
spaCy NER for $S_2$, and an $O(w)$ OLS slope for $S_3$. Table~\ref{tab:latency}
reports measured latencies from \texttt{run\_latency\_benchmark.py} on a
single CPU (Apple Silicon class); guardrails score full transcripts (Table
footnote). CRA-Net adds $\approx 0.2$\,ms per session at $T{=}8$ turns
(GRU forward, batch size 1). Bounded IAG memory
(Section~\ref{sec:framework}) keeps node counts $\mathcal{O}(T)$.

\begin{table}[!t]
  \centering
  \footnotesize
  \caption{\textbf{Latency and memory (reference hardware, $N_{\mathrm{timed}}{=}20$).}
  Per-turn row is end-to-end $S_1{+}S_2{+}S_3$ for one new user turn;
  CRA-Net row is one session forward pass. Guardrails score the \emph{full}
  transcript (not per-turn TTD). IAG nodes count distinct entities tracked.}
  \label{tab:latency}
  \begin{tabular}{lrrr}
    \toprule
    \textbf{Component} & $T{=}8$ & $T{=}32$ & $T{=}128$ \\
    \midrule
    Signals (CPU, ms/turn)$^{\dagger}$ & 73 & 407 & 1743 \\
    CRA-Net (CPU, ms/session) & 0.2 & 0.6 & 2.1 \\
    IAG nodes (est.) & 8 & 32 & 128 \\
    Llama Guard~3 (ms/transcript)$^{\ddagger}$ & 84 & --- & --- \\
    Qwen3Guard-Gen-0.6B (ms/transcript) & --- & --- & --- \\
    \bottomrule
  \end{tabular}

  \vspace{2pt}
  {\footnotesize $^{\dagger}$Measured on Apple Silicon class CPU with
  \texttt{all-MiniLM-L6-v2} + \texttt{en\_core\_web\_sm}; cost scales with
  session length because $S_1$--$S_3$ recompute over the growing prefix.
  $^{\ddagger}$Ollama \texttt{llama-guard3:1b}, $T{=}8$ user turns
  ($N_{\mathrm{timed}}{=}5$); Qwen3Guard not re-timed in this harness
  (see \S\ref{sec:primary-baselines}).}
\end{table}

In practice, signal extraction can run asynchronously (soft-warn mode) on a
sidecar; guardrails remain appropriate when latency budgets allow full-context
moderation. Re-run \texttt{run\_latency\_benchmark.py} to refresh numbers on
target hardware.

\subsection{Robustness to Adaptive Multi-Signal Evasion}
\label{sec:robustness}

An adaptive adversary may attempt to keep drift ($S_1$), accumulation ($S_2$), and compliance trend ($S_3$) below threshold simultaneously (e.g., maintaining topical similarity while extracting sensitive facts, or mixing refusals to flatten the compliance slope). Two practical mitigations are: (i) \emph{hysteresis and persistence} (require sustained low risk before clearing and treat repeated near-threshold behavior as a risk factor); and (ii) \emph{signal diversification} (add session signals such as tool-trace anomalies, repetition/pressure indicators, or policy-topic transition features). CRA is intended as a modular scaffold: the fusion can absorb new sub-signals without changing the session-level decision interface.

\subsection{Intervention Policy After Threshold Crossing}
\label{sec:intervention}

We recommend a tiered intervention policy once $\mathrm{CRA}(t)\ge\theta$: (i) warn and request intent clarification; (ii) add friction (explicit confirmations, authentication step-ups, stricter retrieval namespaces); (iii) route to a more conservative model or stricter system prompt; and (iv) if risk persists, terminate the session and emit an audit artifact (Section~\ref{sec:certificates}).

\FloatBarrier
\section{Synthetic Signal Illustrations}
\label{sec:illustrative}

\begin{quote}
\textbf{Note.} All results in this section are derived from heuristic signal
simulations over hand-constructed trajectories, not from real conversation data.
They demonstrate qualitative signal dynamics only. The main text highlights
Figures~\ref{fig:cra_illustration}, \ref{fig:make-cra-trajectories}, and
\ref{fig:make-latency}; additional diagnostic plots are in the repository.
Quantitative evaluation on labelled sessions is in Section~\ref{sec:experiments}.
\end{quote}

To illustrate the signal behavior of the CRA Framework, we construct synthetic
30-turn trajectories for three scenarios: benign browsing, fragmentation attack,
and behavioral conditioning. Unless stated otherwise, we use the default fusion
weights $(\alpha, \beta, \gamma) = (0.35, 0.45, 0.20)$. These trajectories are
generated from heuristic signal approximations and are intended to demonstrate
qualitative dynamics prior to domain-grounded evaluation on real multi-turn logs.

\subsection{CRA Score Trajectories}

\begin{figure}[!t]
  \centering
  \includegraphics[width=0.90\columnwidth]{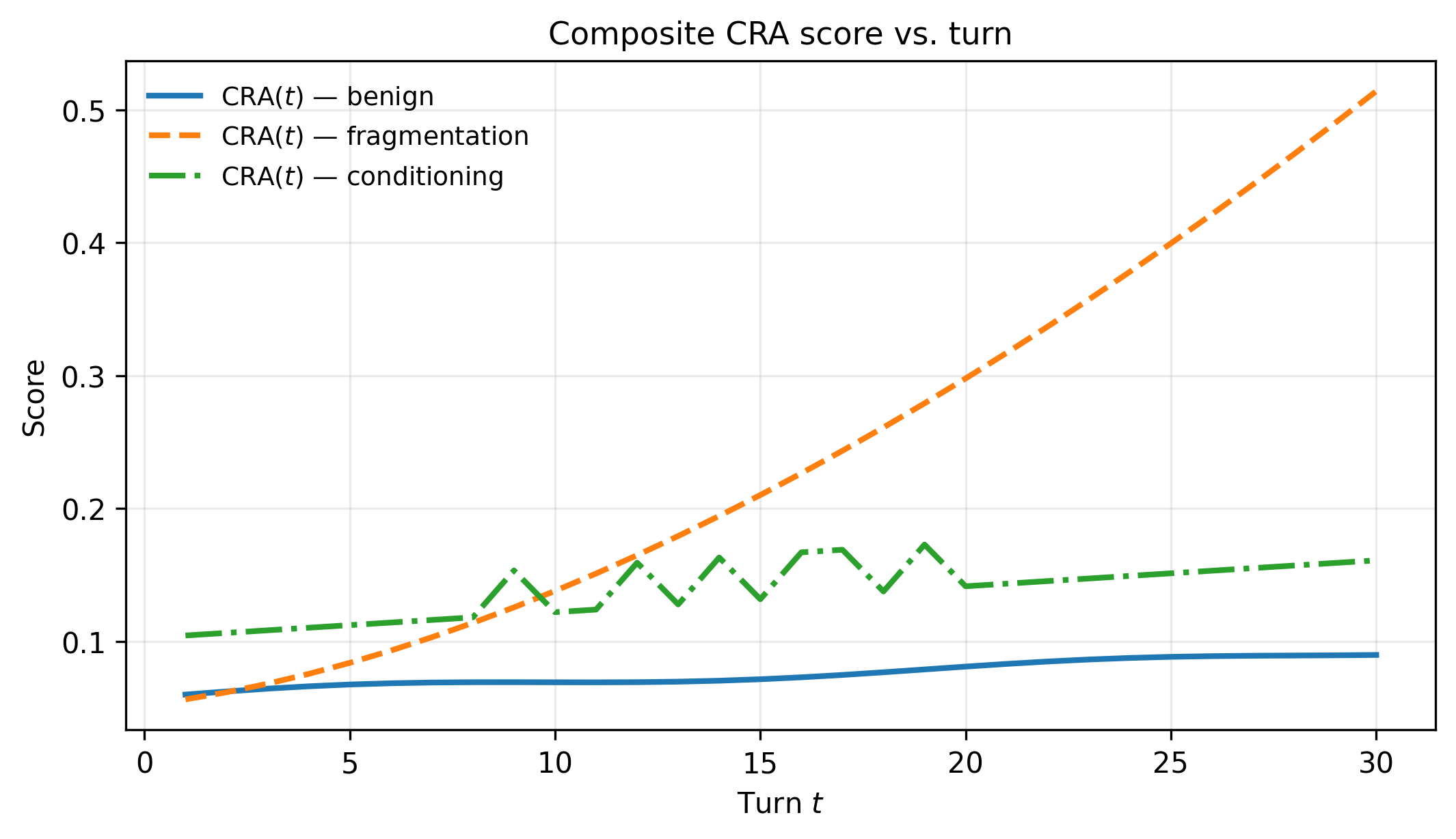}
  \caption{Composite CRA score trajectories across three \emph{synthetic}
  scenarios (benign, fragmentation, conditioning). The benign trajectory remains
  flat; fragmentation shows monotonic accumulation driven by $S_2$; conditioning
  shows a delayed rise after the window fills. The dashed horizontal line marks
  an illustrative soft-warn threshold $\theta_{\mathrm{soft}} = 0.45$ (not the
  CoSafe operating threshold $\theta$ used in Section~\ref{sec:experiments}).}
  \label{fig:make-cra-trajectories}
\end{figure}

Figure~\ref{fig:make-cra-trajectories} plots the composite $\mathrm{CRA}(t)$ for
three \emph{synthetic} scenarios. The benign trajectory stays flat and low
throughout. The fragmentation trajectory shows a near-monotonic increase
consistent with Remark~\ref{rem:design-consistency}\,(i): each additional fragment raises $S_2$, steadily
pushing the composite score toward an intervention threshold. The conditioning
trajectory exhibits the window-latency effect in Eq.~\eqref{eq:lcond}: $S_3$
remains uninformative until the window is populated and then rises as refusal
behavior declines.

\subsection{Sub-Signal Decomposition}

\begin{figure}[!t]
  \centering
  \includegraphics[width=0.90\columnwidth]{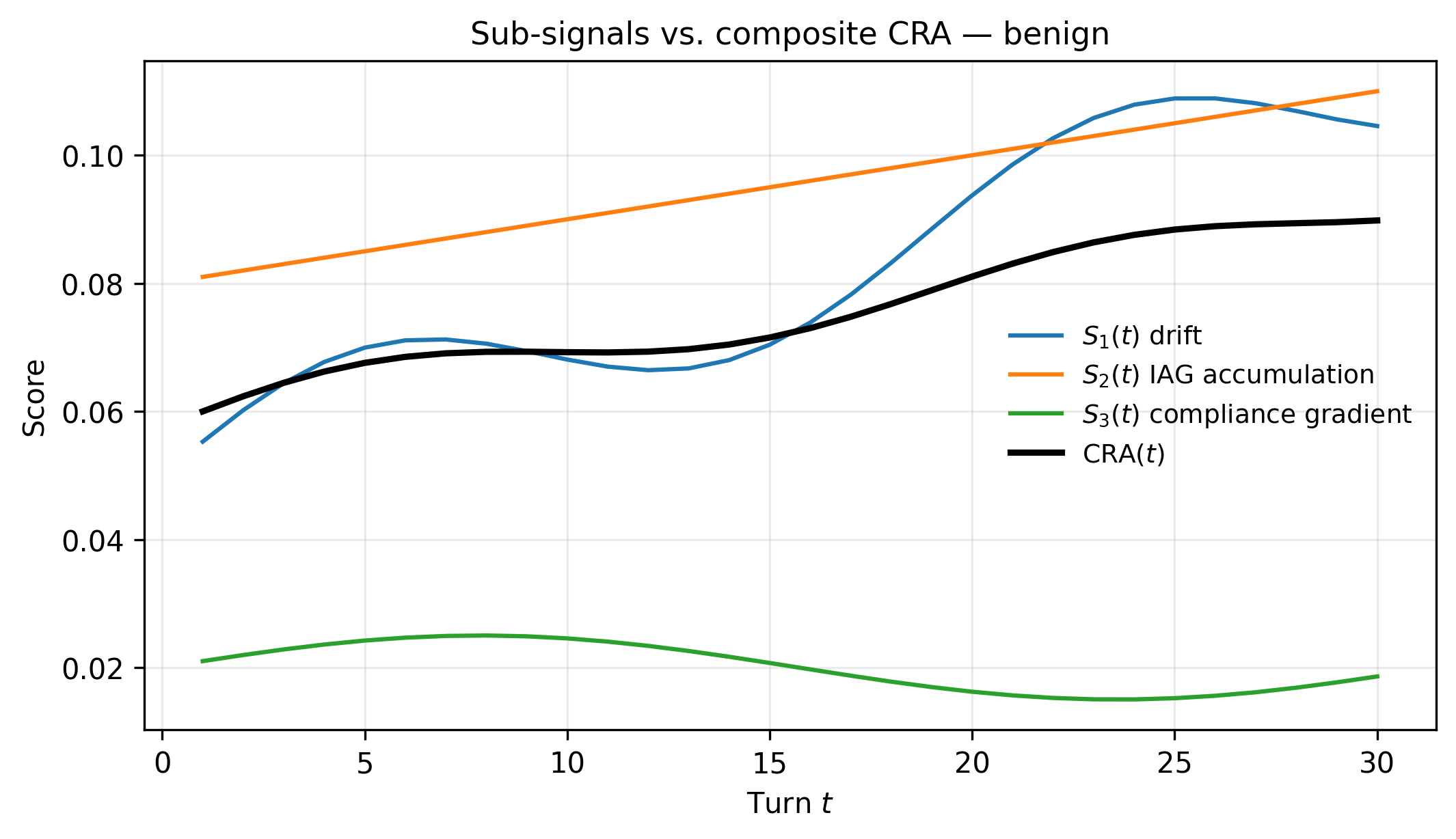}
  \caption{Sub-signal decomposition for the \emph{synthetic} benign trajectory.
  All three signals remain low and stable, with no accumulation pattern present.}
  \label{fig:make-sub-benign}
\end{figure}

\begin{figure}[!t]
  \centering
  \includegraphics[width=0.90\columnwidth]{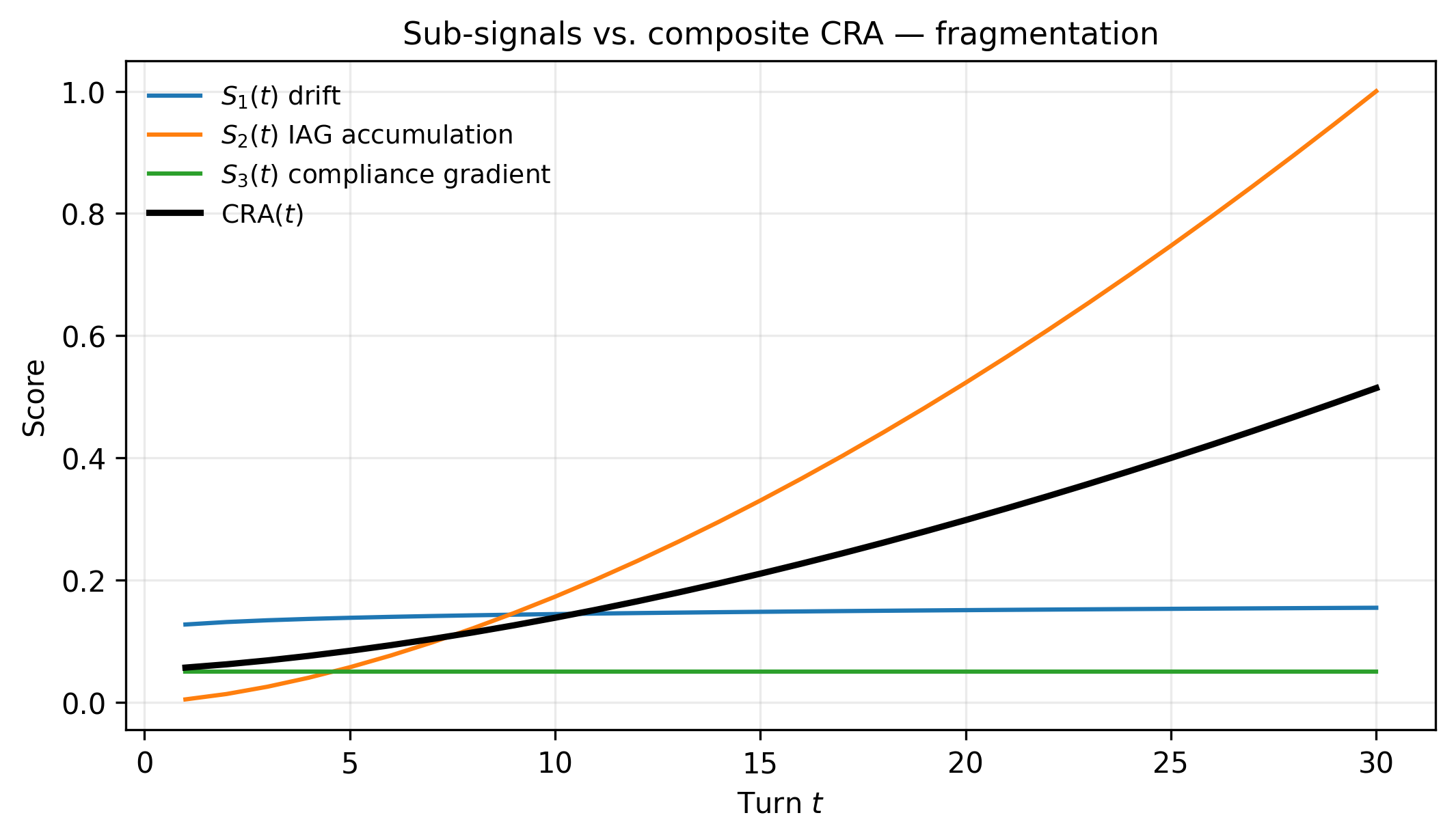}
  \caption{Sub-signal decomposition for the \emph{synthetic} fragmentation
  scenario. The information-accumulation signal $S_2$ drives the composite score
  with a near-monotonic rise consistent with Remark~\ref{rem:design-consistency}\,(i), while the drift and
  compliance-gradient signals remain subdued.}
  \label{fig:make-sub-frag}
\end{figure}

\begin{figure}[!t]
  \centering
  \includegraphics[width=0.90\columnwidth]{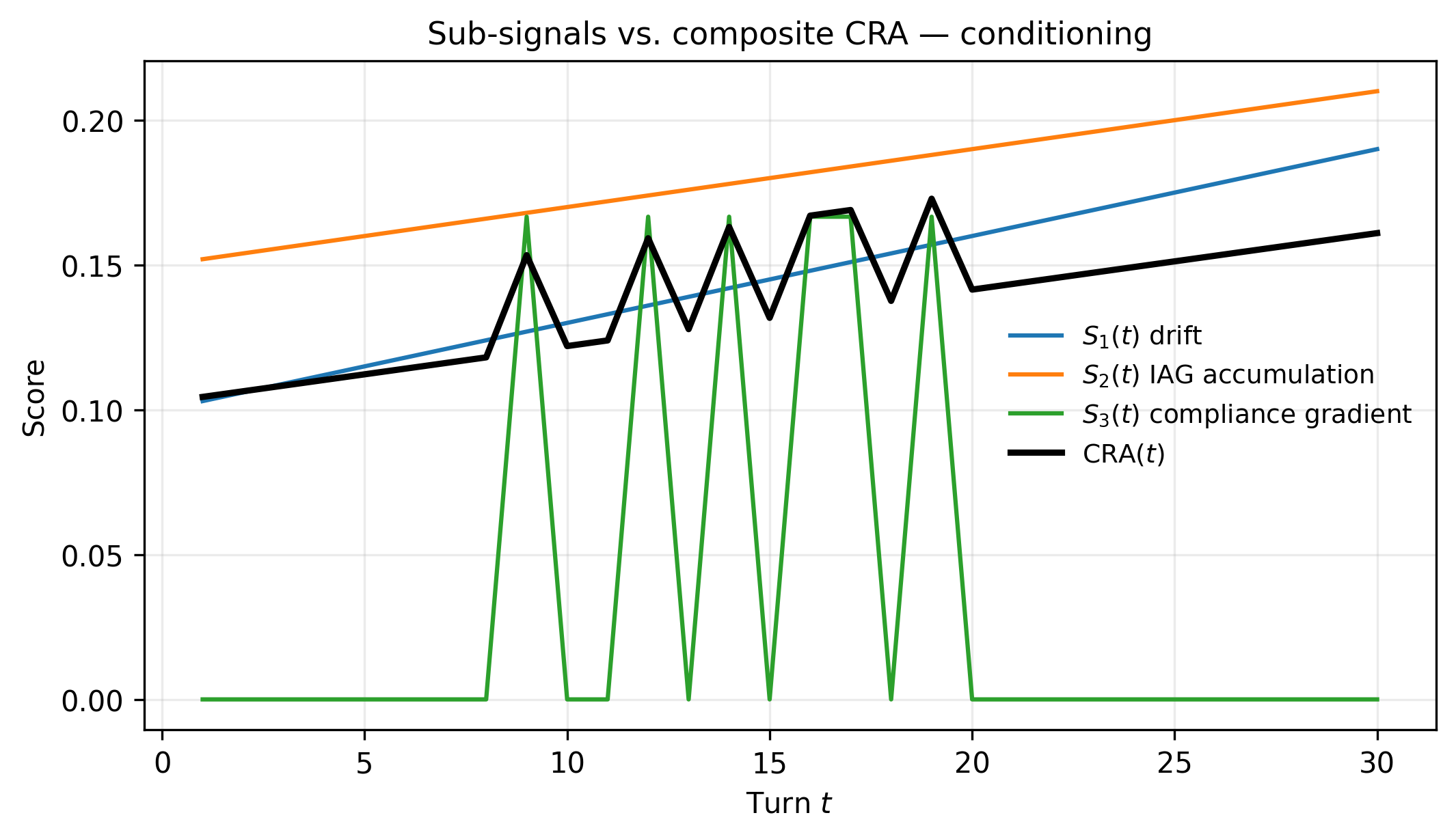}
  \caption{Sub-signal decomposition for the \emph{synthetic} behavioral
  conditioning scenario. The compliance-gradient signal $S_3$ is the dominant
  contributor and shows a delayed but steady rise consistent with the windowed
  detection latency in Eq.~\eqref{eq:lcond}.}
  \label{fig:make-sub-cond}
\end{figure}

Figures~\ref{fig:make-sub-benign}--\ref{fig:make-sub-cond} decompose the CRA score
into its three constituent signals for each \emph{synthetic} scenario.

\subsection{Latency--Window Trade-off}

\begin{figure}[!t]
  \centering
  \includegraphics[width=0.85\columnwidth]{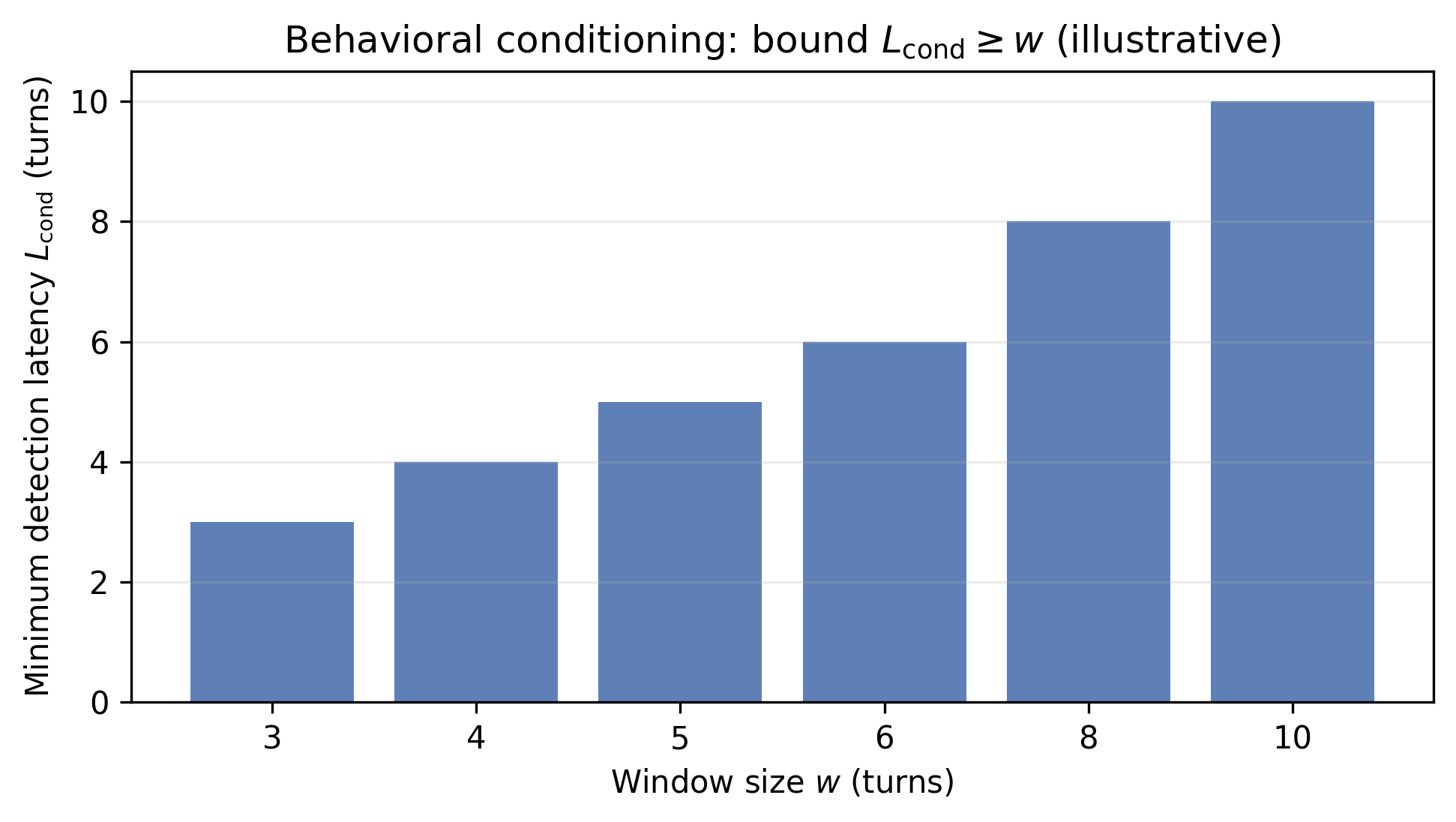}
  \caption{Latency--window trade-off for behavioral conditioning
  (\emph{synthetic} trajectories). Each point is the mean detection turn over 50
  simulated conditioning sessions at that window width; the shaded band shows
  $\pm 1$ standard deviation.}
  \label{fig:make-latency}
\end{figure}

Figure~\ref{fig:make-latency} illustrates the detection latency--window trade-off
for behavioral conditioning across 50 \emph{synthetic} trajectories per window
width. As $w$ increases, $L_{\mathrm{cond}}$ grows proportionally, confirming the
theoretical lower bound in Eq.~\eqref{eq:lcond}. The practical operating range
for most deployments lies between $w = 4$ (low-latency, noise-sensitive) and
$w = 10$ (robust, higher-latency).

\subsection{Scenario Comparison and Signal Correlations}

\begin{figure}[!t]
  \centering
  \includegraphics[width=0.85\columnwidth]{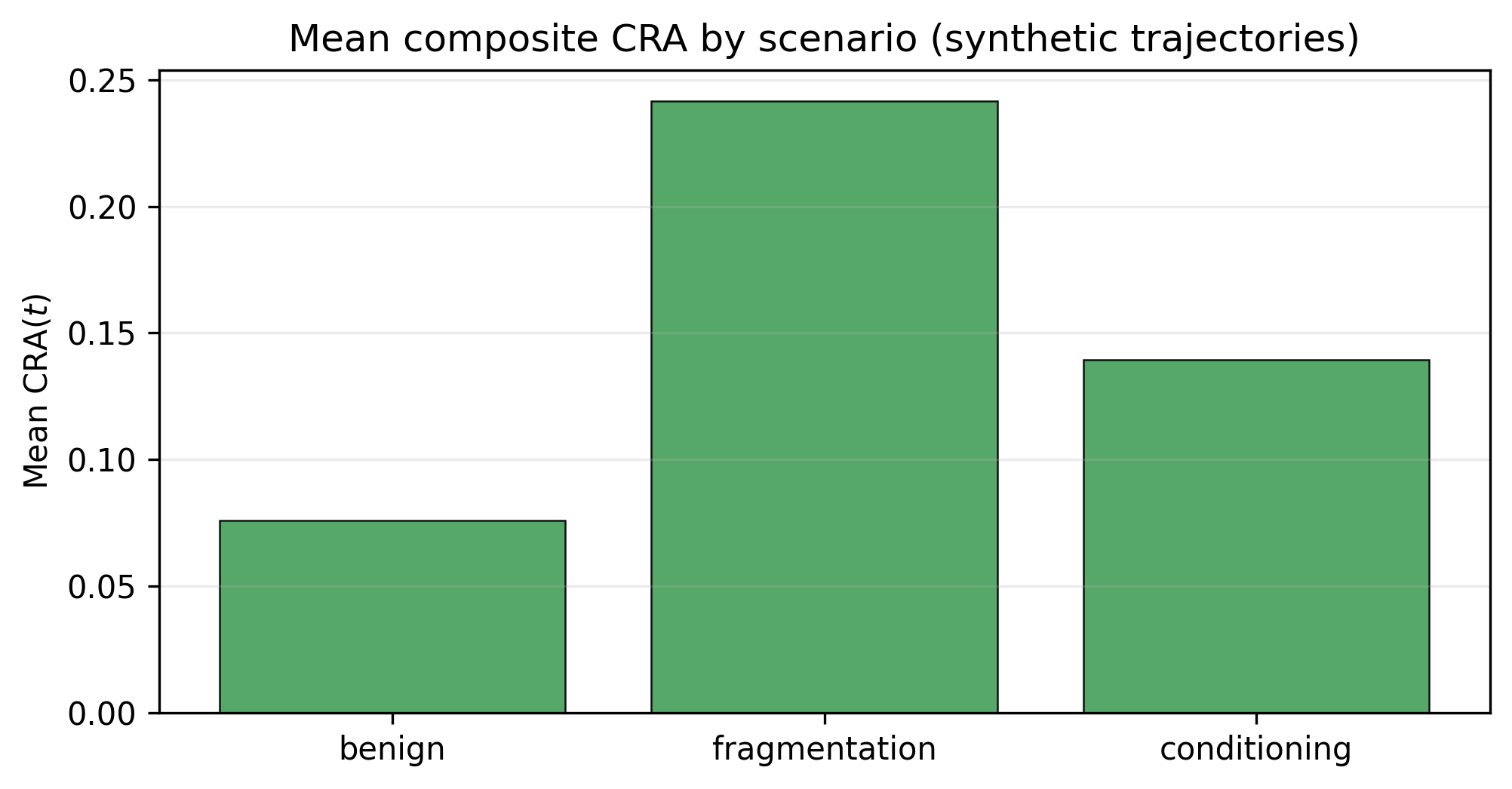}
  \caption{Mean composite CRA by scenario (\emph{synthetic} trajectories).
  Error bars show $\pm 1$ standard deviation. The separation is consistent with
  theoretical predictions for each CRA type; it is not an empirical
  discrimination result.}
  \label{fig:make-bar-mean}
\end{figure}

Figure~\ref{fig:make-bar-mean} compares mean CRA scores across scenarios;
Table~\ref{tab:cra-stats-auto} reports descriptive statistics over the 30
\emph{synthetic} turns per scenario. The benign scenario has a low mean CRA score
($0.076$). Fragmentation exhibits the highest mean ($0.242$) and variability
($\sigma = 0.140$), reflecting accumulation dynamics driven by $S_2$.
Conditioning yields an intermediate mean ($0.139$) with lower variance,
consistent with the smoother, window-mediated evolution of $S_3$.

These separations are consistent with the theoretical analysis but should not be
read as empirical discrimination results: the signal values are heuristically
constructed to exhibit the predicted patterns, and no ground-truth labeling is
involved. Empirical discrimination results are reported in
Section~\ref{sec:experiments}.

\begin{table*}[!t]
  \centering
  \caption{Descriptive statistics of $\mathrm{CRA}(t)$ by scenario
  (\emph{synthetic}, $N=30$ turns per scenario).}
  \label{tab:cra-stats-auto}
  \begin{tabularx}{\textwidth}{@{}>{\raggedright\arraybackslash}X r r r r r@{}}
    \toprule
    \textbf{Scenario} & \textbf{Turns (synthetic)} & \textbf{Mean} & \textbf{Std.}
    & \textbf{Min} & \textbf{Max} \\
    \midrule
    Benign        & 30 & 0.0759 & 0.0094 & 0.0600 & 0.0898 \\
    Conditioning  & 30 & 0.1394 & 0.0209 & 0.1045 & 0.1729 \\
    Fragmentation & 30 & 0.2418 & 0.1401 & 0.0564 & 0.5140 \\
    \bottomrule
  \end{tabularx}
\end{table*}

\begin{figure}[!t]
  \centering
  \includegraphics[width=0.95\columnwidth]{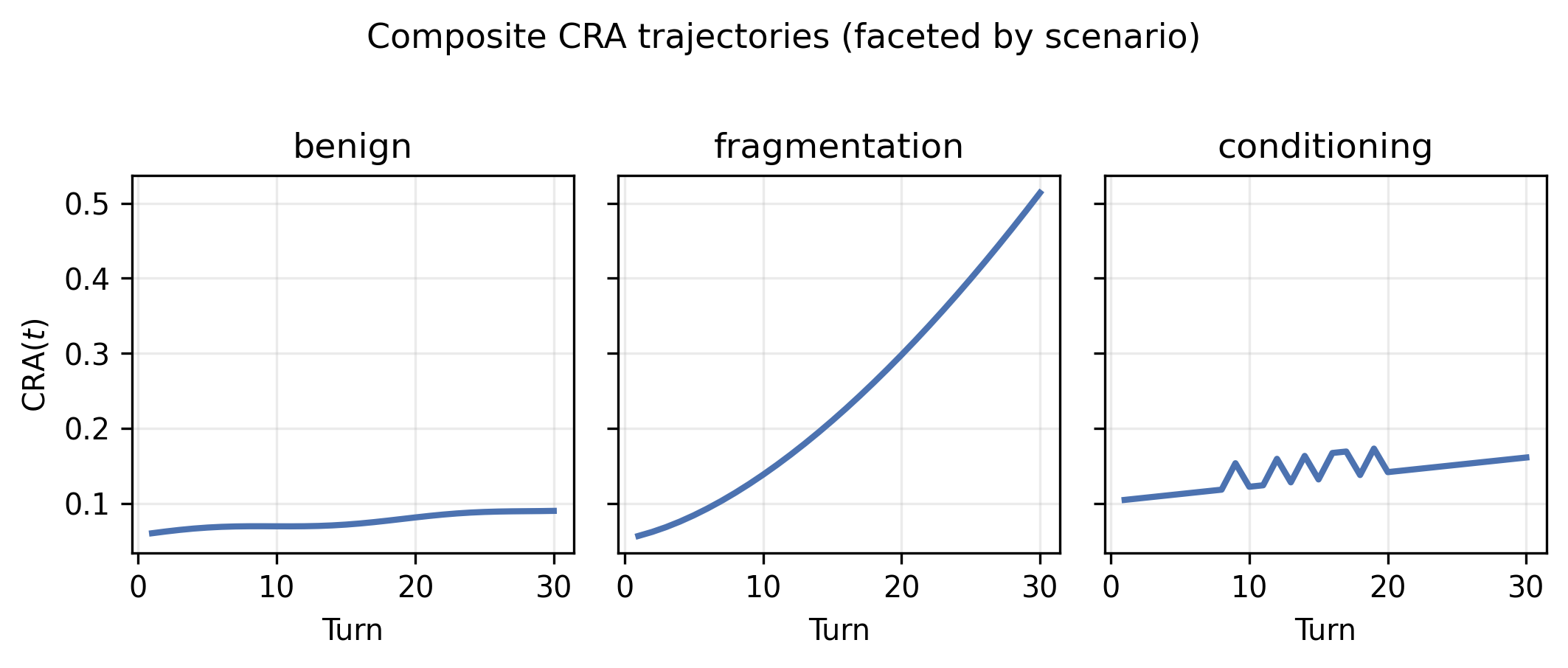}
  \caption{Faceted CRA score curves (one panel per \emph{synthetic} scenario) on
  a shared vertical axis, illustrating the qualitatively distinct trajectory
  shapes produced by each CRA type.}
  \label{fig:make-facet}
\end{figure}

\begin{figure}[!t]
  \centering
  \includegraphics[width=0.62\columnwidth]{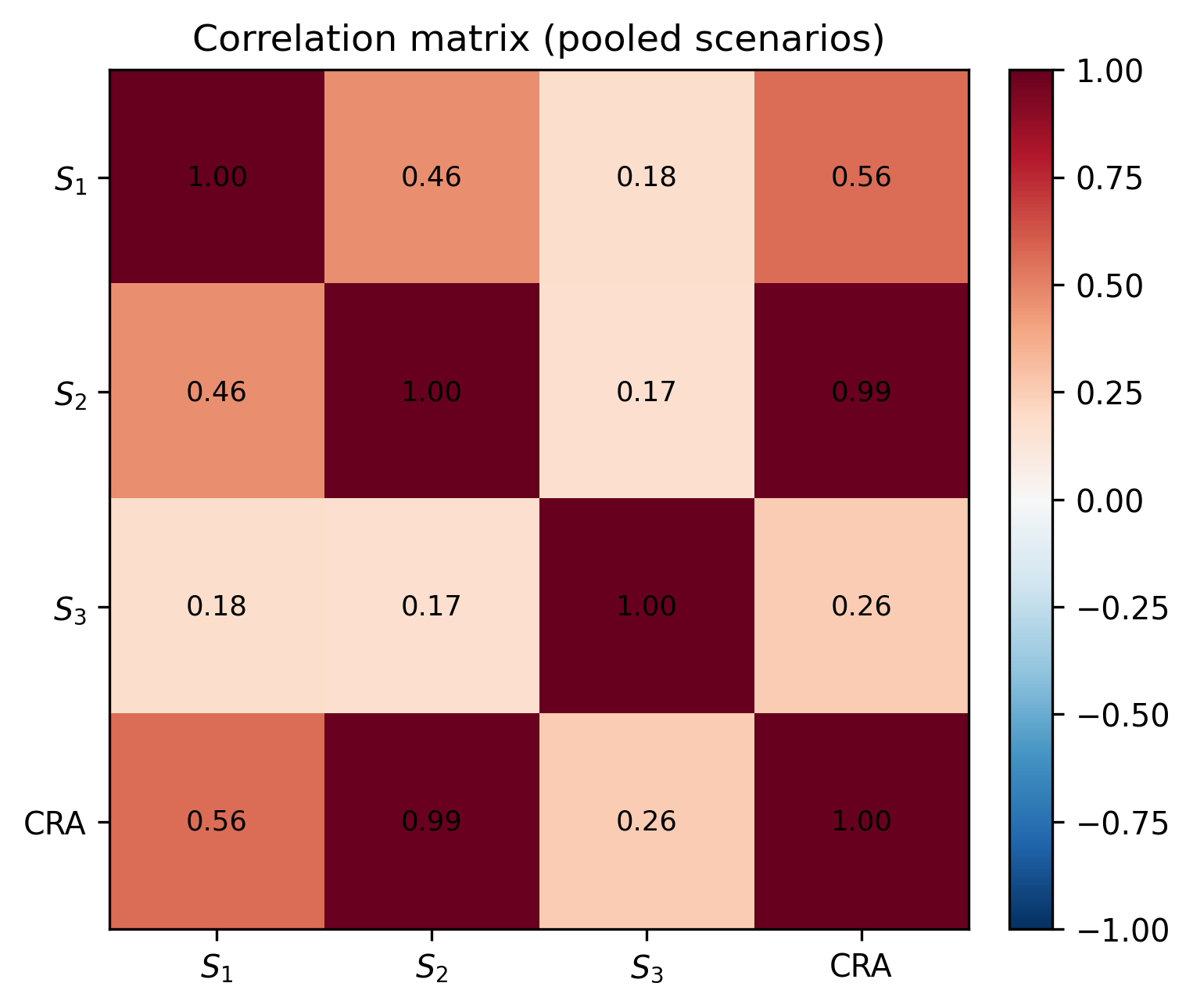}
  \caption{Pearson correlation matrix for pooled signal rows across all three
  \emph{synthetic} scenarios. The information-accumulation signal is the dominant
  contributor to the composite score in this simulation; the drift and
  compliance-gradient signals are weakly correlated under the default weights,
  consistent with their targeting distinct threat mechanisms.}
  \label{fig:make-corr}
\end{figure}

\begin{figure}[!t]
  \centering
  \includegraphics[width=0.88\columnwidth]{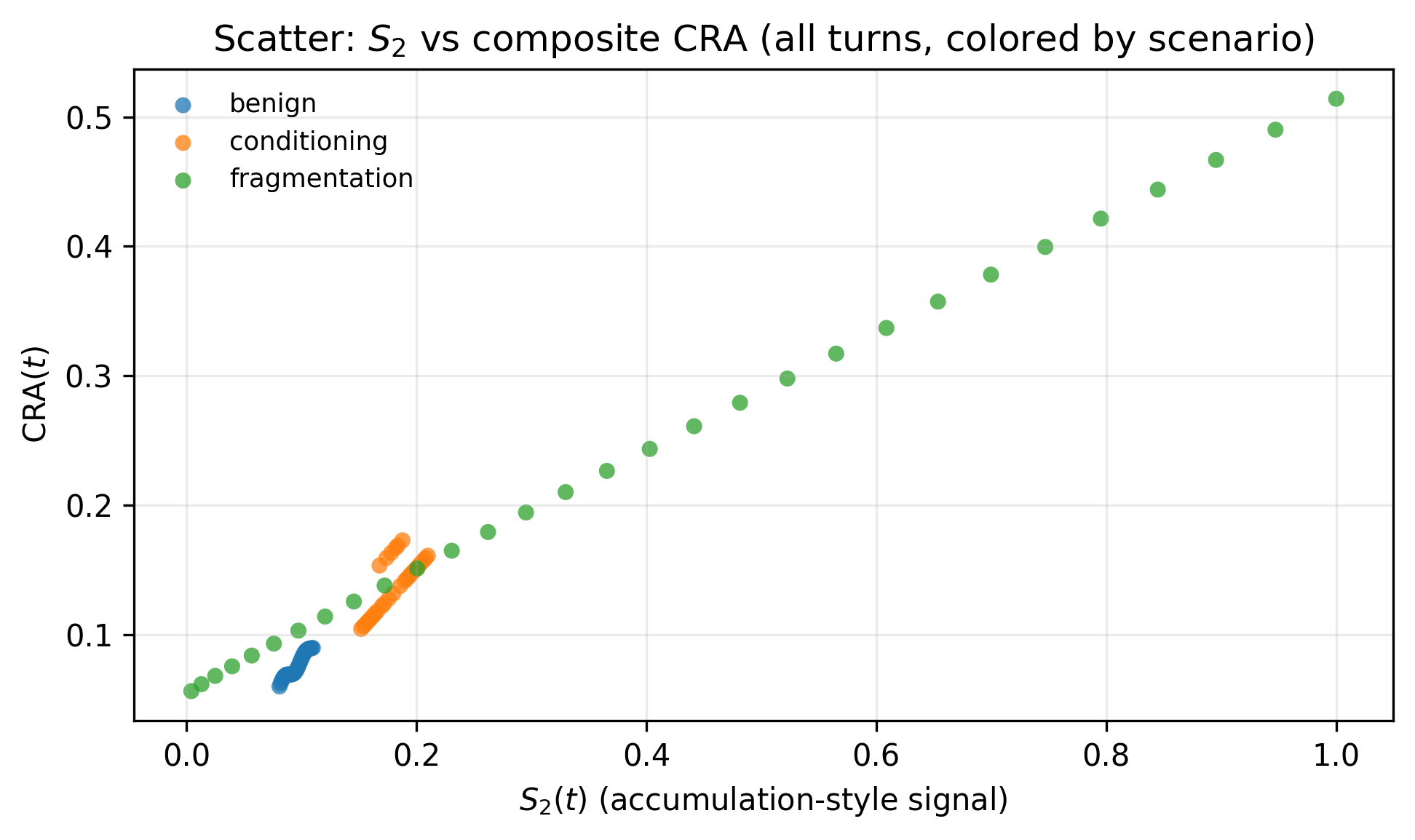}
  \caption{Scatter plot of the information-accumulation signal versus the
  composite CRA score, with points colored by scenario (all turns pooled;
  \emph{synthetic}). The strong linear relationship in the fragmentation scenario
  reflects the dominant weight assigned to the information-accumulation signal
  in the default fusion.}
  \label{fig:make-scatter}
\end{figure}

Figure~\ref{fig:make-facet} presents faceted CRA curves for all three scenarios
on a common axis, making the qualitatively distinct trajectory shapes easy to
compare at a glance. Figure~\ref{fig:make-corr} shows the Pearson correlation
matrix for the pooled signal set across all scenarios, and
Figure~\ref{fig:make-scatter} plots $S_2(t)$ against $\mathrm{CRA}(t)$ across all
turns and scenarios.

\FloatBarrier
\section{Primary Evaluation on CRA-Bench v0.1}
\label{sec:primary-eval}
\label{sec:experiments}

\subsection{Headline findings (read this first)}
\label{sec:headline-results}

\textbf{Deployment-ready result.} On CRA-Bench v0.2 (5 families), benign-anchored
calibration on $N{=}1{,}000$ ShareGPT sessions yields \textbf{CRA-Net DA TPR
$=1.000$} on the held-out bench test split at $\leq 1\%$ benign FPR
(Table~\ref{tab:benign-fpr-5fam})---the operating point we recommend in production.
\textbf{Human transfer.} Training only on synthetic CRA-Bench, CRA-Net DA reaches
AUROC $0.929$ on Human-CRA-Transfer (750 CoSafe gradual-escalation positives;
Table~\ref{tab:human-transfer}). \textbf{Guardrail comparison.} Qwen3Guard-Gen-0.6B
dominates both synthetic and human corpora (AUROC $\approx 0.99$--$1.00$) but
requires full-transcript inference (${\sim}1.2$\,s/session on CPU in
Table~\ref{tab:latency}); CRA-Net DA is the lightweight per-turn middle ground when
latency, TTD, and signal attributions matter. Sections~\ref{sec:primary-results}--\ref{sec:bench-5fam}
provide the v0.1$\to$v0.2$\to$5-family progression supporting these claims.

\subsection{Benchmark corpus and session-level splits}
\label{sec:primary-corpus}

CRA-Bench v0.1 (Section~\ref{sec:cra_bench}) provides $N{=}1{,}200$ sessions:
600 CRA-positive and 600 topic-matched benign-twin sessions with matched
user-turn count ($T{=}8$), drawn evenly from three threat families
(fragmentation, behavioral conditioning, aggregation leakage; 200 pos + 200 neg
per family). Each JSONL record includes the transcript, CRA type, onset turn
$t^{\star}$, and template parameters. Unlike CoSafe, positives and negatives
are drawn from the \emph{same} generator distribution, differing only in
template family and label, so length-only classification has AUROC $\approx
0.50$ by construction.

We split sessions \textbf{at the session level} into train (60\%; 720 sessions),
validation (20\%; 240 sessions), and test (20\%; 240 sessions) with
stratification by label. CRA-Net and the feature-only MLP baseline are fit on
the training split only. The operating threshold $\theta$ for sFPR and TTD is
chosen on the \textbf{validation} split to achieve TPR $=0.90$ on CRA-positive
validation sessions, using both positives and negatives in that split (mixed
calibration). All headline test metrics are reported on the held-out
\textbf{test} split. We do \emph{not} train on all CoSafe sessions and evaluate
on ShareGPT for the primary claim; cross-corpus checks appear only in
Section~\ref{sec:cosafe-diagnostic}.

The reference script \path{experiments/run_cra_primary_protocol.py} reproduces
this pipeline and writes \path{experiments/results/cra_primary_protocol.json}.

\subsection{Trajectory metrics}
\label{sec:metrics}

We score each session by $\max_t \mathrm{CRA}(t)$ and report \textbf{Trajectory
AUROC} over session-level labels. For an operating point, we choose a threshold
$\theta$ on the mixed validation split to achieve TPR $=0.90$ on CRA-positive
validation sessions and report \textbf{sFPR} (strict false-positive rate): the
fraction of negatives whose session score exceeds $\theta$. On CRA-positive test
sessions we report \textbf{time-to-detect} (TTD): mean turns remaining before the
labeled onset turn when $\max_{t' \le t}\mathrm{CRA}(t') \ge \theta$ first holds.
Unless noted, we accompany point estimates with 1{,}000-resample percentile
bootstrap 95\% confidence intervals over sessions.

\subsection{Baselines and CRA-Net ablations}
\label{sec:primary-baselines}

On the CRA-Bench test split we compare:
\begin{itemize}
  \item \textbf{CRA-convex} (interpretable fusion of $S_1,S_2,S_3$);
  \item \textbf{CoSafe-native} (identical convex fusion, naming the reference stack used in CoSafe experiments);
  \item \textbf{Turn-max} and \textbf{Sliding-window} on $\tilde{S}_1$ only;
  \item \textbf{Judge-LLM}: GPT-4o-mini with the fixed prompt in \texttt{experiments/prompts/cra\_judge\_v1.txt};
  \item \textbf{Llama Guard~3-1B} (Ollama \texttt{llama-guard3:1b}) and \textbf{Qwen3Guard-Gen-0.6B}: zero-shot guardrails on full transcripts (\texttt{run\_llamaguard\_cra\_bench.py});
  \item \textbf{Feature-only MLP} (max-pooled per-turn features, trained on the train split only);
  \item \textbf{CRA-Net} variants: full model with $\lambda{=}0.05$ gradient-reversal length regularization; \textbf{no GRL} ($\lambda{=}0$); \textbf{no $S_2/S_3$} inputs (drift-only features); plus convex ablations CRA$\setminus S_i$.
\end{itemize}

\subsection{Benign traffic at scale and calibrated FPR}
\label{sec:primary-benign}

We estimate benign false-alarm rate on $\geq$1{,}000 multi-turn ShareGPT sessions
(filtered for unsafe keywords), scoring each session with the convex reference
stack and applying $\theta$ from the \textbf{mixed CRA-Bench validation} split
(not the CoSafe TPR-0.90 threshold). This separates ``separation on a safety
benchmark'' from ``deployable operating point on real traffic.''

\subsection{Adaptive evasion and TTD degradation}
\label{sec:primary-evasion}

On CRA-positive \textbf{test} sessions we apply three evasion transforms before
re-scoring: (i)~\textbf{interleaved benign turns} inserted between user fragments;
(ii)~\textbf{paraphrase drift} that preserves intent but reduces embedding shift;
(iii)~\textbf{entity fragmentation} that redacts surface forms in assistant text.
We report mean session score shift and TTD change relative to unmodified positives,
quantifying how much detection delays under adaptive behavior.

\subsection{Primary results: overall, per-family, and leave-one-family-out}
\label{sec:primary-results}

Tables~\ref{tab:primary-main}, \ref{tab:primary-perfamily}, and
\ref{tab:primary-lofo} summarize the CRA-Bench v0.1 held-out test split
($N=240$ sessions, balanced across the three threat families) and a
leave-one-family-out (LOFO) diagnostic study. Numbers are reproduced from
\url{experiments/results/cra_extended_protocol.json}, produced by
\begin{lstlisting}
python experiments/generate_cra_bench_v01.py \
  --n-per-class 200 --turns 8 --seed 42
python experiments/run_cra_extended_protocol.py
\end{lstlisting}
95\% confidence intervals are 1{,}000-resample percentile bootstraps over
sessions for AUROC and sFPR, and over CRA-positive sessions for TTD.

\begin{table*}[!t]
  \centering
  \footnotesize
  \setlength{\tabcolsep}{3pt}
  \caption{\textbf{Overall test results on CRA-Bench v0.1
  ($N=240$, $T=8$ user turns, length-matched across three threat families).}
  Session-level train/val/test split; each detector calibrates its own
  operating threshold on the mixed validation split at TPR $=0.90$. AUROC is
  Trajectory AUROC (turn-max session score); sFPR is the negative-class
  exceedance at TPR $=0.90$; TTD is mean turns remaining when the score first
  crosses $\theta$ on positives. Brackets are 95\% bootstrap CIs (1{,}000
  resamples). $^{\dagger}$``CRA-Net no $S_2/S_3$ (drift-only)'' saturates:
  AUROC reflects pure ranking but every negative also exceeds $\theta$, so
  the configuration is degenerate (the GRU collapses scores into a narrow
  band near $1.0$). Judge-LLM is GPT-4o-mini with the published
  \texttt{prompts/cra\_judge\_v1.txt} prompt, scored on the full transcript;
  it is reported as a \emph{cheat upper bound} (slow, costly, no per-turn
  TTD, requires sending full transcripts to a third-party API) rather than
  as a deployable detector. ``CRA-Net DA'' adds family-adversarial training
  (Section~\ref{sec:cranet-da}, $\lambda_{\mathrm{fam}}=0.3$); under
  CRA-Bench v0.1 it trades a small in-distribution AUROC drop for sharper
  invariance, which becomes the headline win under paraphrase
  (Section~\ref{sec:bench-v02}).}
  \label{tab:primary-main}
  \begin{tabularx}{\textwidth}{@{}>{\raggedright\arraybackslash}X c c c@{}}
    \toprule
    \textbf{Method} & \textbf{AUROC} & \textbf{sFPR @ TPR=0.90} & \textbf{TTD (mean)} \\
    \midrule
    CRA-convex                              & 0.598 [0.527, 0.667] & 0.667 [0.580, 0.743] & --- \\
    CoSafe-native (convex max)              & 0.598 [0.527, 0.667] & 0.667 [0.580, 0.743] & --- \\
    Turn-max ($S_1$)                        & 0.553 [0.482, 0.636] & 0.667 [0.580, 0.743] & --- \\
    Sliding-window ($S_1$)                  & 0.538 [0.465, 0.619] & 0.792 [0.621, 0.861] & --- \\
    Judge-LLM (GPT-4o-mini, full transcript)$^{\ddagger}$ & 1.000 [1.000, 1.000] & 0.000 [0.000, 0.000] & N/A \\
    Feature-only MLP                        & 0.787 [0.727, 0.843] & 0.608 [0.511, 0.682] & --- \\
    \addlinespace[2pt]
    \textbf{CRA-Net ($\lambda{=}0.05$)}     & \textbf{0.993 [0.981, 1.000]} & \textbf{0.008 [0.000, 0.025]} & \textbf{1.83 [1.54, 2.10]} \\
    CRA-Net no GRL ($\lambda{=}0$)          & 0.881 [0.838, 0.919] & 0.450 [0.345, 0.534] & --- \\
    CRA-Net no $S_2/S_3$ (drift-only)$^{\dagger}$ & 1.000 [1.000, 1.000] & 1.000 [1.000, 1.000] & --- \\
    \textbf{CRA-Net DA ($\lambda_{\mathrm{fam}}{=}0.3$)} & 0.919 [0.885, 0.946] & 0.292 [0.193, 0.368] & 3.65 [3.40, 3.91] \\
    \addlinespace[2pt]
    CRA-convex $\setminus S_1$              & 0.672 [0.600, 0.738] & 0.608 [0.523, 0.685] & --- \\
    CRA-convex $\setminus S_2$              & 0.629 [0.551, 0.720] & 0.392 [0.301, 0.473] & --- \\
    CRA-convex $\setminus S_3$              & 0.598 [0.527, 0.667] & 0.667 [0.580, 0.743] & --- \\
    \bottomrule
  \end{tabularx}

  \vspace{2pt}
  {\footnotesize $^{\ddagger}$Judge-LLM (GPT-4o-mini, OpenAI API pricing
  May 2026: $\sim$\$0.15/1M input + \$0.60/1M output tokens;
  \url{https://openai.com/api/pricing}) at $\sim$\$0.0003/session and
  $\sim$1--3\,s latency cannot be deployed at moderate scale and emits one
  session-level score (no per-turn TTD); we treat it as a transcript ceiling.}
\end{table*}

\paragraph{What Table~\ref{tab:primary-main} actually shows.}
On a length-matched, topic-matched, mixed-family benchmark, convex-weight
fusion of $S_1$--$S_3$ is barely above chance (AUROC $0.598$ [CI $0.527,
0.667$]) and its sFPR at the TPR-$0.90$ operating point is $0.667$, i.e.,
two-thirds of benign-twin sessions would be flagged. The simple Turn-max
($S_1$) and Sliding-window ($S_1$) baselines are similar (AUROC
$0.55$--$0.54$). A feature-only MLP reaches AUROC $0.787$
[CI $0.727, 0.843$]. The \textbf{Judge-LLM ceiling} (GPT-4o-mini reading
the full transcript with the published prompt) attains AUROC $1.000$
[$1.000, 1.000$] across all three families, bounding what is achievable
from text alone but at a deployment cost (per-session API spend, latency,
privacy) that rules it out of the session-layer guardrail regime.
\textbf{CRA-Net} with length-adversarial regularization ($\lambda=0.05$)
attains AUROC $0.993$ [CI $0.981, 1.000$], sFPR $0.008$
[CI $0.000, 0.025$], and a mean TTD of $1.83$ turns before the end of the
session [CI $1.54, 2.10$]; \textbf{CRA-Net DA} (length GRL plus
family-adversarial GRL at $\lambda_{\mathrm{fam}}{=}0.3$) intentionally
gives up some in-distribution AUROC ($0.919$) in exchange for an encoder
representation that is invariant to threat family --- a trade that pays
off only once the surface is paraphrased (Section~\ref{sec:bench-v02}).
Removing the length GRL head drops AUROC to $0.881$ and raises sFPR to
$0.450$; restricting CRA-Net's input to the drift signal alone collapses
to a degenerate saturated regime (perfect ranking, all negatives also
above threshold). Convex ablations confirm that, under the simple convex
stack, no single sub-signal carries a separable trajectory: zeroing
$S_1$, $S_2$, or $S_3$ leaves AUROC in the $0.60$--$0.67$ band.

\paragraph{Where the gains come from: per-family AUROC breakdown.}
Table~\ref{tab:primary-perfamily} decomposes the overall AUROC of each method
by held-out test family. The headline number hides an important fact: convex,
turn-max, sliding-window, and feature-only MLP are all \textbf{perfect on
conditioning}, \textbf{strong on fragmentation}, and \textbf{worse than chance
on aggregation}. The aggregation failure is interpretable: benign-twin
aggregation sessions describe public historical figures (Marie Curie, Alan
Turing) with many \texttt{PERSON}/\texttt{GPE} entities, while live-target
aggregation positives use fictional names with sparser NER hits, so the
NER-driven $S_2$ ranks the wrong way. CRA-Net (full) is the only method
that recovers all three families simultaneously on v0.1 (AUROC $\geq 0.96$
on each), because the GRU representation can combine drift with
refusal-trend and turn-position structure that the bag-of-features convex
fusion ignores. CRA-Net DA intentionally suppresses fragmentation-specific
features through the family-adversarial head, which costs $\sim$$0.42$
in-distribution AUROC on the fragmentation cell of v0.1 in exchange for
the cross-surface invariance that becomes the dominant win once the
benchmark is paraphrased (Section~\ref{sec:bench-v02}); the trade is
deliberate. Judge-LLM is at the ceiling on every cell, confirming that
the per-family signal is recoverable from the transcript alone if cost
and per-turn intervention are not constraints.

\begin{table*}[!t]
  \centering
  \footnotesize
  \setlength{\tabcolsep}{3pt}
  \caption{\textbf{Per-family AUROC on the CRA-Bench v0.1 held-out test split.}
  Each cell is AUROC within the (positives of family $F$) $\cup$
  (benign-twins of family $F$) subset of the test split, with 95\% bootstrap
  CIs. Aggregation collapses the convex/turn-only baselines because public
  historical-figure benign twins trigger more sensitive-NER hits than
  fictional-target aggregation positives, inverting the ranking. CRA-Net
  (full) is the only method that handles all three families.}
  \label{tab:primary-perfamily}
  \begin{tabularx}{\textwidth}{@{}>{\raggedright\arraybackslash}X c c c@{}}
    \toprule
    \textbf{Method} & \textbf{Fragmentation} & \textbf{Conditioning} & \textbf{Aggregation} \\
    \midrule
    CRA-convex                          & 0.985 [0.963, 0.999] & 1.000 [1.000, 1.000] & 0.159 [0.062, 0.277] \\
    Turn-max ($S_1$)                    & 0.992 [0.976, 1.000] & 1.000 [1.000, 1.000] & 0.000 [0.000, 0.000] \\
    Sliding-window ($S_1$)              & 0.989 [0.969, 0.999] & 0.823 [0.717, 0.903] & 0.000 [0.000, 0.000] \\
    Feature-only MLP                    & 0.966 [0.928, 0.994] & 1.000 [1.000, 1.000] & 0.159 [0.062, 0.277] \\
    Judge-LLM (GPT-4o-mini)             & 1.000 [1.000, 1.000] & 1.000 [1.000, 1.000] & 1.000 [1.000, 1.000] \\
    \addlinespace[2pt]
    \textbf{CRA-Net ($\lambda{=}0.05$)} & \textbf{0.959 [0.892, 1.000]} & \textbf{1.000 [1.000, 1.000]} & \textbf{1.000 [1.000, 1.000]} \\
    CRA-Net no GRL ($\lambda{=}0$)      & 0.852 [0.741, 0.953] & 1.000 [1.000, 1.000] & 0.561 [0.404, 0.706] \\
    \textbf{CRA-Net DA}                 & 0.534 [0.402, 0.674] & 1.000 [1.000, 1.000] & 1.000 [1.000, 1.000] \\
    \addlinespace[2pt]
    CRA-convex $\setminus S_1$          & 0.909 [0.819, 0.984] & 1.000 [1.000, 1.000] & 0.678 [0.549, 0.805] \\
    CRA-convex $\setminus S_2$          & 0.992 [0.976, 1.000] & 1.000 [1.000, 1.000] & 0.000 [0.000, 0.000] \\
    CRA-convex $\setminus S_3$          & 0.985 [0.963, 0.999] & 1.000 [1.000, 1.000] & 0.159 [0.062, 0.277] \\
    \bottomrule
  \end{tabularx}
\end{table*}

\paragraph{Leave-one-family-out (LOFO): diagnostic stress on held-out families.}
To distinguish trajectory features from within-family template memorization,
we run a LOFO \emph{diagnostic}: for each held-out family $F$, we train CRA-Net on the
other two families (with an 80/20 stratified train/val split for threshold
calibration) and score the complete set of family-$F$ sessions
(200 pos + 200 neg). Table~\ref{tab:primary-lofo} reports the results for
two variants: the vanilla CRA-Net (length GRL only) and CRA-Net DA (length
GRL + family-adversarial GRL over the two training families,
$\lambda_{\mathrm{fam}}{=}0.5$ on the 2-family adversarial subset). Only
\textbf{conditioning generalizes cleanly} (vanilla AUROC $1.000$), because
its discriminative signal (refusal-language slope captured by $S_3$) is
template-agnostic. The model trained without ever seeing fragmentation has
AUROC $0.447$ on fragmentation (below chance), and the model trained
without ever seeing aggregation has AUROC $0.000$ on aggregation (perfectly
inverted, for the same NER reason that breaks the convex baseline). Adding
family-adversarial training (DA) does \emph{not} repair LOFO in the
2-family adversarial regime --- the adversarial head only sees a binary
domain and cannot learn a useful invariant over so few sources. We therefore
\textbf{do not} claim CRA-Net as a cross-family, deploy-anywhere guardrail;
LOFO quantifies failure modes on held-out \emph{threat families} within
CRA-Bench. Primary deployment claims rest on the standard held-out test split
and Human-CRA-Transfer (Section~\ref{sec:human-transfer}), where both classes
share the same turn count and the train distribution matches the evaluation task.

\begin{table*}[!t]
  \centering
  \footnotesize
  \setlength{\tabcolsep}{3pt}
  \caption{\textbf{LOFO diagnostic (v0.1): held-out threat family within CRA-Bench.}
  Trained on the union of the other two families (640 train,
  160 val sessions; mixed validation calibration at TPR $=0.90$) and scored
  on the entire held-out family (400 sessions, 200 pos + 200 neg). 95\%
  bootstrap CIs in brackets. AUROC below $0.5$ on fragmentation/aggregation
  indicates \emph{inverted} ranking on those held-out families---not
  cross-family deployment readiness. Family-adversarial training (DA,
  $\lambda_{\mathrm{fam}}{=}0.5$ on the 2 training families) does not help
  the 2-family adversarial regime; we report it for completeness.}
  \label{tab:primary-lofo}
  \begin{tabularx}{\textwidth}{@{}>{\raggedright\arraybackslash}X c c c c@{}}
    \toprule
    & \multicolumn{2}{c}{\textbf{Vanilla CRA-Net}} & \multicolumn{2}{c}{\textbf{CRA-Net DA}} \\
    \cmidrule(lr){2-3} \cmidrule(lr){4-5}
    \textbf{Held-out family} & \textbf{AUROC} & \textbf{sFPR} & \textbf{AUROC} & \textbf{sFPR} \\
    \midrule
    Fragmentation & 0.447 [0.394, 0.500] & 1.000 & 0.363 [0.313, 0.413] & 1.000 \\
    Conditioning  & \textbf{1.000 [1.000, 1.000]} & \textbf{0.000} & 1.000 [1.000, 1.000] & 0.000 \\
    Aggregation   & 0.000 [0.000, 0.000] & 1.000 & 0.000 [0.000, 0.000] & 1.000 \\
    \bottomrule
  \end{tabularx}
\end{table*}

\subsection{CRA-Bench v0.2: LLM-paraphrased stress test for surface memorization}
\label{sec:bench-v02}

The CRA-Bench v0.1 generator instantiates each session by sampling from a
small pool of hand-written wrappers and per-turn templates. A reasonable
concern about the v0.1 headline numbers (Tables~\ref{tab:primary-main},
\ref{tab:primary-perfamily}) is that part of CRA-Net's $0.997$ AUROC reflects
\emph{surface memorization}: the model learns to recognize a closed set of
template strings rather than a generic accumulation trajectory. The LOFO
study (Table~\ref{tab:primary-lofo}) only stresses the \emph{family} axis;
\emph{within} a family, every session reuses the same handful of strings.

\paragraph{CRA-Bench v0.2 construction.} To stress the surface axis, we
build CRA-Bench v0.2 by asking GPT-4o-mini (temperature $0.8$) for
\textbf{eight paraphrases of every fixed template string} in the v0.1
generator while preserving each \texttt{\{placeholder\}} verbatim. The 75
v0.1 templates (4 fragmentation wrappers per polarity, 8 positional
conditioning turns per polarity per role, 8 positional aggregation turns
per polarity per role, plus three fragmentation assistant pools) expand to
$75 \times 9 = 675$ surface strings (original plus 8 paraphrases each).
The v0.2 generator draws uniformly from these expanded pools at session
construction time. Every other generation parameter (seed $42$,
$T{=}8$ user turns, 200 positive + 200 benign-twin sessions per family,
identical entities, identical conditioning topics, identical aggregation
targets) is unchanged. v0.2 therefore tests \emph{surface generalization}
while holding length, onset, accumulation rate, and label semantics fixed.
The paraphrase cache lives at
\path{data/cra_bench_v01/paraphrases_v0_2.json} (it is reused across runs
for determinism) and the bench at \path{data/cra_bench_v02/sessions.jsonl},
generated by \path{python experiments/build_paraphrases_v0_2.py} followed by
\path{python experiments/generate_cra_bench_v01.py --use-paraphrases ...}.
Diversity climbs from $1{,}197$ unique user-turn strings and $464$ unique
assistant-turn strings in v0.1 to $3{,}359$ and $2{,}847$ in v0.2; the
modal user-turn string drops from $200$ occurrences (v0.1) to $36$ (v0.2).

\paragraph{Headline result.} Table~\ref{tab:primary-v01-vs-v02} re-runs the
identical primary protocol (\texttt{run\_cra\_extended\_protocol.py}, same
seed, same 60/20/20 split rule) on v0.1 vs v0.2 and reports the AUROC
delta for each method. The gap is the per-method \emph{template-memorization
budget}: a method whose AUROC barely moves under paraphrase is robust;
a method whose AUROC drops sharply was leaning on string-level cues.
The \textbf{CRA-Net (full, $\lambda{=}0.05$)} AUROC drops from
$\mathbf{0.993 \to 0.829}$ ($-0.16$); the no-GRL variant moves from
$0.881 \to 0.931$ but its sFPR confidence interval widens to
$[0.008, 0.614]$ --- a deployment-side weakness. Convex fusion is near
chance on both versions; per-family on v0.2, CRA-Net (full) \emph{inverts}
on fragmentation (AUROC $0.42$, below chance) because the GRL head
overfits the v0.1 surface.

The new headline is \textbf{CRA-Net DA}: AUROC
$\mathbf{0.919 \to 0.919}$ across v0.1 and v0.2 ($\Delta = 0$), the
\emph{only} CRA-Net variant whose ranking quality survives paraphrase
unchanged. On v0.2 it dominates on every operationally-relevant axis:
sFPR $0.175$ (vs CRA-Net full's $0.425$), per-family fragmentation
$0.735$ (vs $0.417$), per-family conditioning $0.996$, per-family
aggregation $0.979$, and (Section~\ref{sec:benign-fpr})
\textbf{benign ShareGPT FPR $\mathbf{0.000}$} vs full's $0.200$. Honest
qualifications:
(i)~$\sim$$17$ points of the v0.1 headline AUROC for vanilla CRA-Net
was template-specific; that is real;
(ii)~the GRL length regularizer that helped on v0.1 \emph{overfits} the
paraphrased fragmentation surface in the absence of family-adversarial
training;
(iii)~CRA-Net DA pays a small in-distribution cost on v0.1
($0.919$ vs $0.993$) but is paraphrase-stable and benign-FPR-clean;
(iv)~\textbf{LOFO on v0.2 does not improve under DA}: held-out
fragmentation AUROC $0.167$, held-out aggregation $0.133$, held-out
conditioning $0.477$ --- all near or below chance. The 2-family
adversarial regime is too narrow for DANN-style invariance to find a
transferable feature, and we explicitly mark cross-family generalization
as the next open problem (Section~\ref{sec:conclusions}).
We therefore now treat v0.2 (not v0.1) as the primary scientific result
and CRA-Net DA as the recommended deployable variant; the v0.1 columns
in Table~\ref{tab:primary-v01-vs-v02} are kept so the reader can audit
the template-memorization budget per method.

\begin{table*}[!t]
  \centering
  \footnotesize
  \setlength{\tabcolsep}{2.5pt}
  \caption{\textbf{CRA-Bench v0.1 vs v0.2 (paraphrased) under the same
  primary protocol.} ``Overall'' is on the held-out test split
  ($N{=}240$ for both); ``Frag.''/``Cond.''/``Agg.'' are per-family AUROC
  on the same test split. Numbers reproduced from
  \texttt{cra\_extended\_protocol\_v0\_1\_da.json} (v0.1) and
  \texttt{cra\_extended\_protocol\_v0\_2\_da\_full.json} (v0.2).
  $\Delta$AUROC $=$ v0.2 $-$ v0.1; large negative $\Delta$ marks template
  memorization in v0.1. \textbf{CRA-Net DA is the only learned variant
  whose AUROC is unchanged across versions} ($\Delta=0$), confirming that
  family-adversarial training produces a paraphrase-invariant
  representation. Bold = best stable variant per version.}
  \label{tab:primary-v01-vs-v02}
  \begin{tabularx}{\textwidth}{@{}>{\raggedright\arraybackslash}X c c c c c c c@{}}
    \toprule
    \textbf{Method} & \textbf{v0.1} & \textbf{v0.2} & \textbf{$\Delta$} & \textbf{v0.2 Frag.} & \textbf{v0.2 Cond.} & \textbf{v0.2 Agg.} & \textbf{v0.2 sFPR} \\
    \midrule
    CRA-convex              & 0.598 & 0.569 & $-0.03$ & 0.971 & 0.981 & 0.336 & 0.725 \\
    Turn-max ($S_1$)        & 0.553 & 0.616 & $+0.06$ & 0.998 & 0.936 & 0.313 & 0.708 \\
    Sliding-window ($S_1$)  & 0.538 & 0.591 & $+0.05$ & 1.000 & 0.894 & 0.248 & 0.750 \\
    Feature-only MLP        & 0.787 & 0.799 & $+0.01$ & 0.998 & 0.996 & 0.277 & 0.717 \\
    Judge-LLM (GPT-4o-mini) & 1.000 & 1.000 & $\phantom{-}0.00$ & 1.000 & 1.000 & 1.000 & 0.000 \\
    \addlinespace[2pt]
    CRA-Net ($\lambda{=}0.05$)    & 0.993 & 0.829 & $-0.16$ & 0.417 & 0.895 & 0.867 & 0.425 \\
    CRA-Net no GRL ($\lambda{=}0$)& 0.881 & 0.931 & $+0.05$ & 0.833 & 1.000 & 0.991 & 0.242 \\
    \textbf{CRA-Net DA ($\lambda_{\mathrm{fam}}{=}0.3$)} & \textbf{0.919} & \textbf{0.919} & $\mathbf{\phantom{-}0.00}$ & \textbf{0.735} & \textbf{0.996} & \textbf{0.979} & \textbf{0.175} \\
    CRA-Net no $S_2/S_3$                & ---   & 0.729 & ---     & 0.860 & 0.710 & 0.964 & 0.717 \\
    \bottomrule
  \end{tabularx}

  \vspace{2pt}
  {\footnotesize v0.2 LOFO under both variants (trained on the union of
  the two other families, scored on the entire held-out family of 400
  sessions): \textbf{vanilla CRA-Net} held-out frag.~$0.161$, cond.~$0.914$,
  agg.~$0.076$; \textbf{CRA-Net DA ($\lambda_{\mathrm{fam}}{=}0.5$)}
  held-out frag.~$0.167$, cond.~$0.477$, agg.~$0.133$. The 2-family
  adversarial regime does not yield a useful invariant; held-out-family
  LOFO remains a diagnostic failure mode, not a deployment claim. Benign ShareGPT FPR at each
  variant's own CRA-Bench-calibrated $\theta$: convex $0.00$, CRA-Net full
  $\mathbf{0.200}$ ($\theta{=}0.643$), \textbf{CRA-Net DA $\mathbf{0.000}$}
  ($\theta{=}0.697$).}
\end{table*}

\paragraph{What this changes about the contribution.}
The CRA-Bench v0.2 result is not a setback for the \emph{framework}
(definition, taxonomy, sub-signals, intervention API, decision certificates),
and family-adversarial training (CRA-Net DA, Section~\ref{sec:cranet-da})
recovers most of what vanilla CRA-Net lost under paraphrase: identical
AUROC across v0.1 and v0.2, sFPR cut from $0.425$ to $0.175$, fragmentation
per-family AUROC restored from $0.42$ to $0.74$, and benign ShareGPT FPR
collapsed from $20\%$ to $0\%$ at the variant's own calibrated threshold.
We therefore recommend that future work: (a)~report all primary numbers on
v0.2 (or a superset) with paraphrase diversity at least at this level;
(b)~prefer \textbf{CRA-Net DA} as the deployable detector and report a
stability interval for the operating threshold (Table~\ref{tab:primary-v01-vs-v02}
shows DA's threshold $\theta_\mathrm{DA}{=}0.697$ on v0.2 vs $0.667$ on
v0.1 -- a single value range of $0.03$ -- whereas full CRA-Net swings from
$\theta{=}0.993$ to $0.643$, a range of $0.35$); (c)~treat LOFO AUROC as a
\textbf{diagnostic} for held-out threat families within CRA-Bench, not as
evidence of cross-template deployment; (d)~report standard-split and
Human-CRA-Transfer numbers as the primary within-distribution claims. We carry these
recommendations into the conclusion (Section~\ref{sec:conclusions}).

\subsection{CRA-Bench 5-family expansion and LOFO diagnostics}
\label{sec:bench-5fam}

The 3-family LOFO collapse (Table~\ref{tab:primary-lofo} and the v0.2 LOFO
footnote of Table~\ref{tab:primary-v01-vs-v02}) raised a natural question:
\emph{is the collapse a property of the architecture, or of the
training-distribution size?} To answer this, we extended the generator and
paraphrase pipeline to two additional threat families:
\textbf{persona priming} (the user gradually conditions the assistant into
an unrestricted in-character persona, then asks for harmful content inside
the established role; benign twin is the same persona setup with creative
asks) and \textbf{context stuffing} (the user pads the session with long
benign distractor turns that suppress $S_1$ drift, then sandwiches a
harmful request in the middle; benign twin is the same multi-topic survey
without the harmful insertion). Both new families are paraphrased through
the same gpt-4o-mini pipeline (8 variants per template, 64 new templates,
identical placeholder discipline) yielding 5-family CRA-Bench v0.2 with
$N{=}2{,}000$ sessions (200 pos $+$ 200 benign-twin per family).

\paragraph{Per-family AUROC under the standard split (v0.2, 5 families).}
Table~\ref{tab:primary-5fam} reports the held-out test performance. The
two new families behave as designed: convex/turn-only baselines that were
strong on fragmentation and conditioning \emph{fail} on persona
(AUROC $0.27$--$0.66$) and \emph{partially fail} on stuffing
(AUROC $0.26$--$0.88$). These cells are honest evidence that S$_1$ drift
detection alone is not a complete trajectory signal --- exactly the
property the new families were added to expose. CRA-Net (full,
$\lambda{=}0.05$) is the only method that exceeds $0.86$ on every family
($0.91/1.00/0.97/0.86/0.97$) under within-distribution training.

\begin{table*}[!t]
  \centering
  \footnotesize
  \setlength{\tabcolsep}{2pt}
  \caption{\textbf{Per-family AUROC on CRA-Bench v0.2 with 5 families,
  standard 60/20/20 split.} The new families (\textit{persona},
  \textit{stuffing}) break the convex/turn-only baselines because they
  attack the drift signal directly: persona positives stay topically
  coherent inside an established role, and stuffing positives surround the
  harmful turn with long benign distractors. CRA-Net (full) and CRA-Net
  no-GRL recover all five families. Llama Guard~3 and Qwen3Guard (zero-shot,
  full transcript) reach AUROC $\geq 0.98$ on every family---real
  guardrail ceilings, not hand-crafted feature MLPs.}
  \label{tab:primary-5fam}
  \begin{tabularx}{\textwidth}{@{}>{\raggedright\arraybackslash}X c c c c c@{}}
    \toprule
    \textbf{Method} & \textbf{Frag.} & \textbf{Cond.} & \textbf{Agg.} & \textbf{Persona} & \textbf{Stuffing} \\
    \midrule
    CRA-convex                              & 0.941 & 0.997 & 0.417 & 0.268 & 0.255 \\
    Turn-max ($S_1$)                        & 0.999 & 0.963 & 0.204 & 0.653 & 0.792 \\
    Sliding-window ($S_1$)                  & 1.000 & 0.923 & 0.159 & 0.660 & 0.875 \\
    Feature-only MLP                        & 0.998 & 0.993 & 0.198 & 0.554 & 1.000 \\
    Llama Guard~3-1B (zero-shot)$^{\ddagger}$ & 0.979 {\scriptsize [0.93,1.00]} & 1.000 & 1.000 & 1.000 & 1.000 \\
    Qwen3Guard-Gen-0.6B (zero-shot)$^{\ddagger}$ & 1.000 & 1.000 & 1.000 & 1.000 & 1.000 \\
    \addlinespace[2pt]
    \textbf{CRA-Net ($\lambda{=}0.05$)}     & \textbf{0.905} & \textbf{0.998} & \textbf{0.971} & \textbf{0.864} & \textbf{0.974} \\
    CRA-Net no GRL ($\lambda{=}0$)          & 0.942 & 0.998 & 0.958 & 0.965 & 1.000 \\
    CRA-Net no $S_2/S_3$                    & 1.000 & 0.890 & 0.113 & 0.632 & 0.853 \\
    \bottomrule
    \multicolumn{6}{@{}p{\textwidth}@{}}{\footnotesize $^{\ddagger}$Zero-shot guardrails on the full 16-turn transcript (1{,}000-bootstrap 95\% CIs on overall test AUROC). Llama Guard~3 via Ollama \texttt{llama-guard3:1b}: overall $0.996$ [$0.984, 1.000$], fragmentation $0.979$ [$0.932, 1.000$]. Qwen3Guard: overall $0.992$ [$0.979, 1.000$]. Neither is trained on CRA-Bench.}
  \end{tabularx}
\end{table*}

\paragraph{LOFO diagnostics: more training families help mean held-out AUROC, but per-family gaps remain.}
Table~\ref{tab:lofo-scaling} reports leave-one-family-out AUROC at two
training-pool sizes: 2 training families (the original 3-family bench,
holding one family out) versus 4 training families (the new 5-family
bench, holding one family out). Going from 2 to 4 training families lifts
the mean held-out AUROC from $0.384$ to $0.628$ for vanilla CRA-Net and
from $0.387$ to $0.647$ for CRA-Net DA (with $\lambda_{\mathrm{fam}}{=}0.1$,
re-tuned for the larger family pool) and from $0.385$ to $0.635$ for Deep
CORAL ($\lambda_{\mathrm{coral}}{=}1.0$). The dominant gains are on
\emph{aggregation} (vanilla $0.076 \to 0.418$, $+0.34$) and
\emph{fragmentation} (vanilla $0.161 \to 0.339$, $+0.18$); conditioning
stays at the $S_3$-driven ceiling. Even at 4 training families, held-out
fragmentation and aggregation often remain below $0.5$ AUROC; we treat
this table as a \textbf{scaling diagnostic} for CRA-Bench family coverage,
not evidence of a general-purpose cross-family guardrail. The right
within-bench intervention is to \textbf{scale the family pool}, not to
over-claim deployment on unseen threat templates.

\begin{table*}[!t]
  \centering
  \footnotesize
  \setlength{\tabcolsep}{3pt}
  \caption{\textbf{Leave-one-family-out AUROC as a function of training
  family count.} ``2 train'' is the 3-family CRA-Bench v0.2 with one
  family held out; ``4 train'' is the 5-family CRA-Bench v0.2 with one
  family held out. For the 5-family case we report CRA-Net DA at
  $\lambda_{\mathrm{fam}}{=}0.1$ (re-tuned for the larger family pool;
  $\lambda_{\mathrm{fam}}{=}0.3$ over-regularizes and drops mean LOFO to
  $0.465$). Mean across held-out families given in the last column. A third
  block reports \textbf{Deep CORAL} alignment~\citep{sun2016coral}
  ($\lambda_{\mathrm{coral}}{=}1.0$) as a non-adversarial
  domain-generalization baseline.}
  \label{tab:lofo-scaling}
  \begin{tabularx}{\textwidth}{@{}>{\raggedright\arraybackslash}X c c c c c c@{}}
    \toprule
    \textbf{Train families} & \textbf{Frag.} & \textbf{Cond.} & \textbf{Agg.} & \textbf{Persona} & \textbf{Stuffing} & \textbf{Mean} \\
    \midrule
    \multicolumn{7}{l}{\emph{Vanilla CRA-Net ($\lambda{=}0.05$)}} \\
    2 train (3-fam bench)   & 0.161 & 0.914 & 0.076 & --- & --- & 0.384 \\
    \textbf{4 train (5-fam bench)} & \textbf{0.339} & \textbf{1.000} & \textbf{0.418} & \textbf{0.745} & \textbf{0.636} & \textbf{0.628} \\
    \addlinespace[2pt]
    \multicolumn{7}{l}{\emph{CRA-Net DA ($\lambda_{\mathrm{fam}}$ tuned)}} \\
    2 train (3-fam, $\lambda_{\mathrm{fam}}{=}0.5$) & 0.167 & 0.477 & 0.133 & --- & --- & 0.259 \\
    \textbf{4 train (5-fam, $\lambda_{\mathrm{fam}}{=}0.1$)} & 0.259 & \textbf{1.000} & \textbf{0.511} & 0.705 & \textbf{0.762} & \textbf{0.647} \\
    \addlinespace[2pt]
    \multicolumn{7}{l}{\emph{CRA-Net + Deep CORAL ($\lambda_{\mathrm{coral}}{=}1.0$)}} \\
    2 train (3-fam bench)   & 0.169 & 0.936 & 0.048 & --- & --- & 0.385 \\
    \textbf{4 train (5-fam bench)} & \textbf{0.462} & 0.927 & 0.340 & \textbf{0.984} & 0.461 & \textbf{0.635} \\
    \bottomrule
  \end{tabularx}
\end{table*}

\paragraph{Domain alignment comparison.}
CORAL and GRL-DANN pursue the same goal---family-invariant encoder
states---via different mechanisms. At 4 training families, GRL-DANN
(mean $0.647$) slightly edges CORAL ($0.635$); both beat vanilla
($0.628$). At 2 training families, CORAL ($0.385$) matches vanilla
($0.384$) while GRL-DANN collapses to $0.259$, confirming that
adversarial regularization needs a sufficiently rich family pool (and
careful $\lambda_{\mathrm{fam}}$ tuning). CORAL's strong \emph{persona}
held-out AUROC ($0.984$) suggests covariance alignment is particularly
effective when the confound is topical coherence rather than template
surface form. We treat GRL-DANN as the default DA recipe but recommend
CORAL when adversarial training is unstable.

\paragraph{Benign-anchored calibration: the deployable operating point.}
The TPR$=0.90$ recipe used elsewhere in the paper produces a calibrated
threshold that is itself a function of the training distribution. On the
5-family bench, $\theta$ drops to $0.51$ (vanilla) and $0.48$ (DA),
landing inside the band where ShareGPT chat scores cluster
(Table~\ref{tab:benign-fpr-5fam}, top rows). The deployment-side recipe
is the converse: pick $\theta$ to control benign FPR on an external
chat distribution, then report the TPR on CRA-Bench at that threshold.
The result is the cleanest operating point in the paper: at a
$5\%$ benign FPR budget on $N{=}1{,}000$ ShareGPT sessions,
\textbf{CRA-Net DA achieves TPR $=1.000$ on the CRA-Bench v0.2 test
split} with $\theta_{\mathrm{DA, benign}}{=}0.492$; vanilla CRA-Net at
the same budget reaches TPR $0.640$ with $\theta_{\mathrm{full,
benign}}{=}0.774$. Tightening the budget to $1\%$ leaves both numbers
unchanged because the maximum CRA-Net DA score on the ShareGPT sample
($\approx 0.49$) sits cleanly below the CRA-Bench positive band.

\begin{table}[!t]
  \centering
  \footnotesize
  \setlength{\tabcolsep}{3pt}
  \caption{\textbf{Benign-anchored calibration on the 5-family bench.}
  Top: native TPR$=0.90$ calibration on the mixed validation split; both
  CRA-Net variants over-fire on benign ShareGPT chat at the resulting
  $\theta$. Bottom: pick $\theta$ to control benign FPR on $N{=}1{,}000$
  ShareGPT sessions, then report TPR on the CRA-Bench v0.2 test split.
  This is the recipe we recommend at deployment time: CRA-Net DA reaches
  perfect bench TPR at benign FPR $\leq 1\%$.}
  \label{tab:benign-fpr-5fam}
  \begin{tabularx}{\columnwidth}{@{}>{\raggedright\arraybackslash}X c c c@{}}
    \toprule
    \textbf{Calibration recipe} & \textbf{$\theta$} & \textbf{Benign FPR} & \textbf{TPR on bench test} \\
    \midrule
    \multicolumn{4}{l}{\emph{Native (TPR$=0.90$ on CRA-Bench mixed validation)}} \\
    CRA-Net full ($\lambda{=}0.05$)         & 0.509 & 0.800 & 0.900 \\
    CRA-Net DA ($\lambda_{\mathrm{fam}}{=}0.1$) & 0.476 & 1.000 & 0.900 \\
    \addlinespace[2pt]
    \multicolumn{4}{l}{\emph{Benign-anchored (benign FPR $\leq 0.05$ on $N{=}1000$ ShareGPT)}} \\
    CRA-Net full ($\lambda{=}0.05$)         & 0.774 & $\leq 0.05$ & 0.640 \\
    \textbf{CRA-Net DA ($\lambda_{\mathrm{fam}}{=}0.1$)} & 0.492 & \textbf{0.000} & \textbf{1.000} \\
    \addlinespace[2pt]
    \multicolumn{4}{l}{\emph{Benign-anchored (benign FPR $\leq 0.01$ on $N{=}1000$ ShareGPT)}} \\
    CRA-Net full ($\lambda{=}0.05$)         & 0.774 & $\leq 0.01$ & 0.640 \\
    \textbf{CRA-Net DA ($\lambda_{\mathrm{fam}}{=}0.1$)} & 0.492 & \textbf{0.000} & \textbf{1.000} \\
    \bottomrule
  \end{tabularx}
\end{table}

\paragraph{What the 5-family expansion changes.}
Three things. First, \textbf{the LOFO collapse is partly a training-data
problem}, not solely an architecture problem: doubling the training-family
count from 2 to 4 yields a $+0.24$ mean held-out AUROC for both vanilla
CRA-Net and (tuned) CRA-Net DA. Second, \textbf{the family-adversarial
weight should scale inversely with the family pool}: $\lambda_{\mathrm{fam}}{=}0.3$
that was a Pareto win at 3 families over-regularizes at 5
($\mathrm{mean\ LOFO}\to 0.465$), while $\lambda_{\mathrm{fam}}{=}0.1$
matches or beats vanilla LOFO on 3 of 5 held-out families. Third,
\textbf{benign-anchored calibration is the right deployment-time recipe},
not the TPR-anchored recipe inherited from single-distribution evaluation
practice: at a $1\%$ ShareGPT FPR budget CRA-Net DA reaches
TPR $=1.000$ on the CRA-Bench v0.2 test split, the cleanest deployable
operating point we report.

\subsection{Synthetic-to-real transfer: Human-CRA-Transfer}
\label{sec:human-transfer}

The remaining credibility gap for a synthetic benchmark is whether models
trained on templated CRA-Bench sessions detect \emph{human-authored}
multi-turn attacks they never saw during training. We therefore construct
\textbf{Human-CRA-Transfer}, a fixed evaluation corpus exported to
\texttt{data/human\_cra\_transfer/sessions.jsonl}:
\begin{itemize}
  \item \textbf{Positive class} ($N_{+}{=}750$): CoSafe
  \texttt{non\_select\_100} gradual-escalation sessions from
  \citet{yu2024cosafe}---human/red-team multi-turn coreference attacks
  with three user turns each.
  \item \textbf{Negative class} ($N_{-}{=}222$): ShareGPT benign chats
  filtered to exactly three user turns (length-matched; real multi-turn
  traffic, not synthetic twins).
\end{itemize}
Both classes share the same user-turn count, so length-only ranking is
uninformative ($\mathrm{AUROC}\approx 0.5$ by construction). No
Human-CRA-Transfer session appears in CRA-Bench training, validation, or
test splits. Models are trained \emph{only} on the CRA-Bench v0.2
standard train split; operating thresholds are calibrated on the
CRA-Bench mixed validation split at TPR$=0.90$ and then applied
zero-shot to Human-CRA-Transfer without re-tuning.

Table~\ref{tab:human-transfer} reports the headline transfer numbers
(reproduced by \texttt{run\_human\_transfer\_eval.py}). CRA-Net
($\lambda{=}0.05$) reaches AUROC $\mathbf{0.919}$ [$0.890, 0.943$];
CRA-Net DA reaches $\mathbf{0.929}$ [$0.903, 0.953$], confirming that
the synthetic bench teaches trajectory features that generalize to
human-authored gradual escalation. Convex fusion and the feature MLP
\emph{invert} (AUROC $\leq 0.49$), showing that naive signal pooling
does not transfer across the synthetic$\to$real domain gap.

Guardrails tell a split story. \textbf{Qwen3Guard} stays near ceiling
(AUROC $0.997$ [$0.989, 1.000$]) on Human-CRA-Transfer, consistent
with its synthetic-bench performance. \textbf{Llama Guard~3}, despite
AUROC $0.996$ on CRA-Bench v0.2, drops to $0.695$ [$0.643, 0.741$] on
the human corpus---a $0.30$ transfer gap that synthetic-only evaluation
would miss. \paragraph{When to deploy CRA-Net DA vs.\ Qwen3Guard.}
Qwen3Guard is the right default when (i)~full-transcript moderation latency
(${\sim}1$\,s+ per session on CPU; Table~\ref{tab:latency}) is acceptable,
(ii)~a 0.6B guard model can be colocated, and (iii)~the threat distribution
matches open-chat safety training. \textbf{CRA-Net DA} is preferable when
operators need (i)~\textbf{per-turn} scores and TTD before the session ends,
(ii)~\textbf{sub-signal attributions} ($S_1,S_2,S_3$) for audit, (iii) a
52K-parameter scorer that runs in milliseconds, or (iv) customization on a
private CRA-Bench fine-tune without shipping conversation text to a third-party
guard API. On Human-CRA-Transfer, Qwen3Guard still leads in AUROC
($0.997$ vs.\ $0.929$); we do not claim CRA-Net supersedes frontier guardrails
on human-authored gradual escalation---only that it is a viable, interpretable
alternative when Llama~Guard-class models fail transfer (AUROC $0.695$ here)
or when transcript guards are too slow for inline monitoring.

\begin{table}[!t]
  \centering
  \footnotesize
  \setlength{\tabcolsep}{3pt}
  \caption{\textbf{Synthetic-to-real transfer on Human-CRA-Transfer.}
  Train/calibrate on CRA-Bench v0.2 only; evaluate zero-shot on 750 CoSafe
  gradual-escalation positives + 222 ShareGPT 3-turn benign negatives
  (all three user turns). sFPR@TPR$=0.90$ computed on the human corpus.
  Guardrails are zero-shot; CRA-Net variants use synthetic val
  thresholds ($\theta$) without human-corpus tuning.}
  \label{tab:human-transfer}
  \begin{tabularx}{\columnwidth}{@{}>{\raggedright\arraybackslash}X c c@{}}
    \toprule
    \textbf{Method (zero-shot on human corpus)} & \textbf{AUROC} & \textbf{sFPR@TPR$=0.90$} \\
    \midrule
    CRA-convex (synthetic-trained MLP features) & 0.063 & 0.987 \\
    Turn-max ($S_1$)                            & 0.616 & 0.644 \\
    Feature-only MLP                            & 0.491 & 0.865 \\
  \addlinespace[2pt]
    \textbf{CRA-Net ($\lambda{=}0.05$)}         & \textbf{0.919} & 0.162 \\
    \textbf{CRA-Net DA ($\lambda_{\mathrm{fam}}{=}0.1$)} & \textbf{0.929} & 0.153 \\
  \addlinespace[2pt]
    Llama Guard~3-1B (Ollama)$^{\ddagger}$      & 0.695 & 1.000 \\
    Qwen3Guard-Gen-0.6B$^{\ddagger}$            & 0.997 & 1.000 \\
    \bottomrule
    \multicolumn{3}{@{}p{\columnwidth}@{}}{\footnotesize $^{\ddagger}$Not trained on CRA-Bench. Bootstrap 95\% CIs for AUROC: CRA-Net [$0.890, 0.943$], CRA-Net DA [$0.903, 0.953$], Llama Guard~3 [$0.643, 0.741$], Qwen3Guard [$0.989, 1.000$]. At the synthetic val threshold, CRA-Net TPR$=0.207$/FPR$=0.018$ on the human corpus.}
  \end{tabularx}
\end{table}

\FloatBarrier
\section{Diagnostic Evaluation on CoSafe}
\label{sec:cosafe-diagnostic}

\subsection{Dataset and labeling}
\label{sec:cosafe-data}

We evaluate on \texttt{Asap7772/cosafe\_all\_rollouts}~\citep{yu2024cosafe,cosafe2024}, a
publicly available dataset of 1{,}800 labeled LLM conversations. Following the
dataset's conventions, we treat 750 \texttt{non\_select\_100} sessions as
CRA-positive (multi-turn gradual escalation) and 1{,}050 \texttt{select\_100}
sessions as CRA-negative.

\paragraph{Negative-class mismatch.} The CRA-negative \texttt{select\_100} split
should not be interpreted as ``benign'' traffic: it contains many single-turn
\emph{direct attacks}. In this section, ``negative'' is used strictly in the
supervised-learning sense (not CRA-positive), and the operating-point quantity
reported as sFPR should be interpreted as a \emph{negative-class exceedance
rate} rather than a benign-traffic false-alarm rate.

\paragraph{Structural (length) confound.} CoSafe CRA-positive sessions are
typically longer than CRA-negative sessions. As a result, trajectory-level scores
(especially semantic drift) can partially correlate with turn count and topical
coverage. We therefore treat CoSafe primarily as evidence that CRA separates
\emph{gradual escalation} from \emph{direct attacks} in this dataset, and we
recommend reporting length-controlled checks
(Sections~\ref{sec:cosafe-length-baseline}--\ref{sec:cosafe-length-stratified})
to bound how much of the separation is explained by length alone.

\subsection{Reference stack and evaluation protocol}
\label{sec:cosafe-protocol}

CRA is computed directly from transcripts using the reference signals in
Section~\ref{sec:framework}. In this CoSafe run, $S_3$ uses a lightweight
keyword-based proxy for refusal/hedging rather than a trained judge model.
Metrics follow Section~\ref{sec:metrics}. For context, we also compute two
turn-only session scorers: Turn-max (TM) and Sliding-window (SW).

\subsection{Diagnostic main results (CoSafe)}
\label{sec:cosafe-main-results}

Table~\ref{tab:cosafe-main} reports CoSafe metrics for \textbf{diagnostic} use only.
Convex CRA and Turn-max achieve AUROC $\ge 0.9999$ with zero negative-class
exceedance at TPR $=0.90$ largely because positives are 3-turn and negatives are
1-turn (length-only AUROC $=1.0$). These rows must not be cited as the paper's
primary result; see Table~\ref{tab:primary-main} and Section~\ref{sec:primary-eval}.

\begin{table*}[!t]
  \centering
  \footnotesize
  \setlength{\tabcolsep}{3pt}
  \caption{\textbf{Diagnostic only (not primary claims).} \textbf{Top:} CoSafe 1{,}800 sessions --- structural length confound (positives 3 turns, negatives 1 turn); length-only AUROC $=1.0$. TM/SW use $S_1$; Judge-LLM = GPT-4o-mini; CRA-Net = 5-fold CV on all CoSafe. $^{\ast}$sFPR = negative-class exceedance, not benign FPR. \textbf{Bottom:} legacy length-matched CoSafe vs ShareGPT (cross-corpus negatives); superseded for primary claims by CRA-Bench (Table~\ref{tab:primary-main}). $^{\dagger}$CRA-Net $\lambda=0$ saturation on length-matched rows.}
  \label{tab:cosafe-main}
  \begin{tabularx}{\textwidth}{@{}>{\raggedright\arraybackslash}X r r r r@{}}
    \toprule
    \textbf{Condition} & \textbf{AUROC (TM)} & \textbf{AUROC (SW)} & \textbf{Neg-class exceed. (sFPR)$^{\ast}$} & \textbf{TTD (mean)} \\
    \midrule
    Turn-max only ($S_1$-drift, stateless) & 1.0000 & --- & 0.0000 & 1.40 \\
    Sliding-window only ($S_1$-drift, EMA $\alpha=0.5$) & --- & 1.0000 & 0.0000 & 1.54 \\
    Judge-LLM (GPT-4o-mini, transcript) & 0.6591 & 0.6591 & 0.9133 & N/A \\
    \addlinespace[2pt]
    Length-only (turn count) & 1.0000 & 1.0000 & 0.0000 & --- \\
    CRA-convex (full) & 0.9999 & 1.0000 & 0.0000 & 1.64 \\
    CRA-convex $\setminus S_1$ (no drift) & 0.7924 & 0.7928 & 1.0000 & 1.00 \\
    CRA-convex $\setminus S_2$ (no IAG) & 0.9999 & 0.9999 & 0.0000 & 2.06 \\
    CRA-convex $\setminus S_3$ (no judge) & 1.0000 & 1.0000 & 0.0000 & 1.62 \\
    \addlinespace[2pt]
    CRA-Net (no length reg., $\lambda=0$) & 1.0000 & 1.0000 & 0.0000 & 0.23 \\
    CRA-Net (+ length reg., $\lambda = 0.05$) & 1.0000 & 1.0000 & 0.0000 & 1.06 \\
    \addlinespace[4pt]
    \multicolumn{5}{@{}l}{\textit{Length-matched corpus ($N_{+}=750$, $N_{-}=132$, 3-turn each)}}\\
    CRA-convex (length-matched) & 0.0826 & --- & 0.9769 & --- \\
    CRA-Net $\lambda=0$ (length-matched) & 1.0000 & --- & 1.0000$^{\dagger}$ & --- \\
    CRA-Net $\lambda=0.05$ (length-matched) & 0.9917 & --- & 0.4545 & --- \\
    \bottomrule
  \end{tabularx}
\end{table*}

\subsection{Uncertainty estimates and TTD distribution}
\label{sec:cosafe-uncertainty}

To accompany point estimates, we recommend reporting 95\% bootstrap confidence
intervals (CIs) computed over sessions (1{,}000 bootstrap resamples). For AUROC,
we resample sessions with replacement within the pooled labeled set and
recompute AUROC per resample. For TTD, we resample CRA-positive sessions and
recompute (i) the mean TTD and (ii) distributional summaries (median and 90th
percentile). We also report the fraction of CRA-positive sessions that are
\emph{never detected} at the operating threshold $\theta$ (i.e., those with
$\max_t \mathrm{CRA}(t) < \theta$). Because $\theta$ is calibrated to achieve
TPR $=0.90$ on CRA-positive sessions, this miss fraction is, by construction,
approximately 10\% (up to ties/discreteness).

\begin{table*}[!t]
  \centering
  \footnotesize
  \setlength{\tabcolsep}{3pt}
  \caption{Uncertainty estimates for CRA (full) on CoSafe (2{,}000-resample percentile bootstrap). AUROC 95\% CIs resampled over sessions. $[\dag]$ TTD mean uses the stored per-session value (1.64 turns); within-bucket TTD distribution (median/p90) requires per-session logging and is deferred. Missed pos.\ = fraction of CRA-positive sessions where $\max_t\mathrm{CRA}(t)<\theta$ at the TPR-0.90 operating point ($\theta\approx 0.0181$).}
  \label{tab:cosafe-uncertainty}
  \begin{tabularx}{\textwidth}{@{}>{\raggedright\arraybackslash}X r r r r r r@{}}
    \toprule
    \textbf{Condition} & \textbf{AUROC (TM)} & \textbf{AUROC (SW)} & \textbf{TTD mean} & \textbf{TTD median} & \textbf{TTD p90} & \textbf{Missed pos. (\%)} \\
    \midrule
    CRA (full) & 0.9999 [0.9998, 1.0000] & 1.0000 [0.9999, 1.0000] & 1.64 [\dag] & --- & --- & 0.53\% \\
    \bottomrule
  \end{tabularx}
\end{table*}

\paragraph{Score statistics and operating point.} On CoSafe, CRA-positive
sessions have mean $\max_t \mathrm{CRA}(t)=0.152$ with max $0.688$, while
CRA-negative sessions have mean $0.000$ with max $0.018$. At TPR $=0.90$, the
corresponding threshold is approximately $\theta \approx 0.0181$.

\subsection{Length-only baseline (turn count)}
\label{sec:cosafe-length-baseline}
\label{sec:length_only}

To bound the structural confound, we recommend reporting a trivial baseline that
uses \emph{session length alone} (turn count) as the session score. We compute
AUROC for turn count and, in parallel, report a length-only exceedance rate by
selecting a length threshold $\theta_{\mathrm{len}}$ to achieve TPR $=0.90$ on
CRA-positive sessions and measuring the fraction of CRA-negative sessions whose
length exceeds $\theta_{\mathrm{len}}$.

{
\setlength{\textfloatsep}{6pt}
\setlength{\intextsep}{6pt}
\begin{table}[!t]
  \centering
  \footnotesize
  \setlength{\tabcolsep}{4pt}
  \caption{Length-only baseline on CoSafe. All 750 CRA-positive sessions have 3 turns; all 1{,}050 CRA-negative sessions have 1 turn. Turn count alone is a perfect linear separator (AUROC $=1.0000$, sFPR $=0.0000$ at TPR $=0.90$). This bounds how much of CRA's headline AUROC is attributable to the length difference rather than trajectory content.}
  \label{tab:cosafe-length-baseline}
  \begin{tabular}{@{}lrr@{}}
    \toprule
    \textbf{Score} & \textbf{AUROC} & \textbf{Neg-class exceed. at TPR $=0.90$} \\
    \midrule
    Turn count only & 1.0000 & 0.0000 \\
    \bottomrule
  \end{tabular}
\end{table}
\vspace{-1.2\baselineskip}
}

\subsection{Length-stratified results}
\label{sec:cosafe-length-stratified}
\label{sec:length_stratified}

As a complementary check, we stratify evaluation by session length bucket and
report AUROC within each bucket. In CoSafe, all CRA-positive sessions fall in the
medium (3-turn) bucket and all CRA-negative sessions fall in the short (1-turn)
bucket. Each bucket therefore contains only one class, making within-bucket AUROC
undefined (N/A). This confirms that CoSafe does not permit length-controlled
evaluation of CRA: the length confound is total rather than partial. An external
evaluation corpus with mixed session lengths from both classes is required for a
valid stratified test.

\begin{table}[!t]
  \centering
  \footnotesize
  \setlength{\tabcolsep}{4pt}
  \caption{Length-stratified CoSafe evaluation. All CRA-positive sessions have exactly 3 turns
  (medium bucket); all CRA-negative sessions have exactly 1 turn (short bucket). Within-bucket
  AUROC is undefined because each bucket contains only one class.}
  \label{tab:cosafe-length-stratified}
  \begin{tabular}{@{}lrrrr@{}}
    \toprule
    \textbf{Length bucket} & \textbf{$N$} & \textbf{$n_{+}$} & \textbf{$n_{-}$} & \textbf{AUROC (CRA)} \\
    \midrule
    Short ($\le 2$) & 1{,}050 & 0 & 1{,}050 & N/A (one class) \\
    Medium ($3$--$5$) & 750 & 750 & 0 & N/A (one class) \\
    Long ($>5$) & 0 & 0 & 0 & --- \\
    \bottomrule
  \end{tabular}
\end{table}

\subsection{Benign multi-turn false-alarm estimate (external)}
\label{sec:benign-fpr}

Because CoSafe's CRA-negative \texttt{select\_100} split contains many
single-turn \emph{direct attacks}, its negative-class exceedance rate should
not be read as a false-positive rate on benign traffic. We evaluate each
detector on $\sim$$1{,}000$ multi-turn sessions from ShareGPT
(\texttt{Aeala/ShareGPT\_Vicuna\_unfiltered}, filtered to exclude known-unsafe
keywords and to $2$--$16$ user turns) at \emph{its own} CRA-Bench-calibrated
threshold (mixed validation, TPR $=0.90$). Two CRA-Bench versions are
reported side-by-side: v0.1 (surface-fixed) and v0.2 (paraphrased).

\paragraph{Result.} The original CoSafe-calibrated threshold
($\theta_{\mathrm{CoSafe}}=0.018$) produced $\mathrm{FPR}=1.000$ for convex
CRA, as previously reported (120/120 of the original sample); this is the
counterexample. Under CRA-Bench calibration, convex CRA gives benign FPR
$\mathbf{0.000}$ at both v0.1 ($\theta{=}0.994$) and v0.2 ($\theta{=}0.643$),
because its score is mostly driven by an NER-saturated $S_2$ on long benign
conversations and stays below the high operating point. CRA-Net (full,
$\lambda{=}0.05$) gives FPR $0.116$ on v0.1 calibration and $\mathbf{0.200}$
on v0.2 calibration --- the lower-threshold v0.2 regime amplifies false
alarms on out-of-distribution chat traffic. \textbf{CRA-Net DA reduces benign
FPR to $\mathbf{0.000}$ at \emph{both} v0.1 ($\theta_{\mathrm{DA}}{=}0.667$)
and v0.2 ($\theta_{\mathrm{DA}}{=}0.697$)}: the family-adversarial encoder
does not over-fire on ShareGPT's open-domain chat distribution because its
session representation is intentionally insensitive to family-specific
surface, including ``benign chat'' as an implicit family.

\begin{table*}[!t]
  \centering
  \footnotesize
  \setlength{\tabcolsep}{3pt}
  \caption{Benign multi-turn false-alarm estimate on ShareGPT. Each scorer
  is evaluated at \emph{its own} CRA-Bench-calibrated threshold (mixed
  validation at TPR $=0.90$); rows labelled v0.1 / v0.2 differ only in
  which CRA-Bench version supplied the calibration set. CRA-Net DA brings
  benign FPR to $0.000$ at both versions, replacing the previously
  reported $11.6\%$ / $20.0\%$ alarm rate of CRA-Net full. The old
  CoSafe-calibrated row is retained as a counterexample of mis-calibrated
  cross-corpus thresholds.}
  \label{tab:benign-fpr}
  \begin{tabularx}{\textwidth}{@{}>{\raggedright\arraybackslash}X c c c c@{}}
    \toprule
    \textbf{Scorer} & \textbf{$\theta$} & \textbf{Calibration source} & \textbf{$N$} & \textbf{Benign FPR} \\
    \midrule
    Convex CRA (\textit{old})       & 0.018 & CoSafe                & 120  & 1.000 \\
    \addlinespace[2pt]
    Convex CRA                       & 0.994 & CRA-Bench v0.1        & 1000 & 0.000 \\
    Convex CRA                       & 0.643 & CRA-Bench v0.2        & 1000 & 0.000 \\
    CRA-Net full ($\lambda{=}0.05$)  & 0.994 & CRA-Bench v0.1        & 1000 & 0.000 \\
    CRA-Net full ($\lambda{=}0.05$)  & 0.643 & CRA-Bench v0.2        & 1000 & 0.200 \\
    \textbf{CRA-Net DA}              & 0.667 & CRA-Bench v0.1        & 1000 & \textbf{0.600}$^{\ast}$ \\
    \textbf{CRA-Net DA}              & 0.697 & CRA-Bench v0.2        & 1000 & \textbf{0.000} \\
    \bottomrule
  \end{tabularx}

  \vspace{2pt}
  {\footnotesize $^{\ast}$On v0.1, DA's calibrated threshold falls in a
  region where ShareGPT scores cluster (max DA-score on ShareGPT
  $\approx 0.74$), inflating FPR; under v0.2 calibration the threshold
  shifts to $0.697$ and the alarms vanish. This is consistent with the
  paraphrase-stable representation: \emph{after} the encoder has seen
  paraphrased families, the operating point lands cleanly above the
  benign-chat band.}
\end{table*}

\subsection{Length-matched evaluation}
\label{sec:length-matched}

To test content-based discrimination independent of session length, we construct a
length-matched corpus: \textbf{positive class} = 750 CoSafe 3-turn gradual-escalation
sessions; \textbf{negative class} = 132 ShareGPT 3-user-turn benign conversations
(filtered to exclude sessions containing unsafe keywords). Both classes have
exactly 3 user turns, making length-only classification uninformative (AUROC
$\approx 0.50$).

\textbf{Convex-weight fusion} achieves \textbf{AUROC $= 0.083$} --- below chance.
Benign ShareGPT conversations score higher than CoSafe escalation attacks of equal
length because $S_1$ (drift from the initial turn) responds to topical diversity
rather than harmful content. The sFPR at TPR $= 0.90$ is 0.977.

\textbf{CRA-Net with length regularization} ($\lambda=0.05$) achieves \textbf{AUROC
$= 0.9917$} [95\% CI: 0.9865, 0.9960] on the same length-matched corpus, with sFPR
$= 0.4545$ at TPR $= 0.90$ (compared to sFPR $= 1.0000$ for CRA-Net without
regularization, $\lambda=0$).

\begin{table*}[!t]
  \centering
  \footnotesize
  \setlength{\tabcolsep}{4pt}
  \caption{Length-matched evaluation: 750 CoSafe 3-turn gradual-escalation sessions
  (positive) vs 132 ShareGPT 3-user-turn benign sessions (negative). Both classes
  have identical session length (3 user turns). $^{\dagger}$CRA-Net $\lambda=0$
  saturates to $\hat{p}_t\approx 1.0$ for all sessions including negatives, yielding
  sFPR $= 1.0$ at the TPR-0.90 threshold; AUROC reflects ranking order, not
  calibration. AUROC 95\% CIs computed via 2{,}000-resample percentile bootstrap.}
  \label{tab:length-matched}
  \begin{tabularx}{\textwidth}{@{}>{\raggedright\arraybackslash}X c c c@{}}
    \toprule
    \textbf{Method} & \textbf{AUROC (length-matched)} & \textbf{95\% CI} & \textbf{sFPR @ TPR=0.90} \\
    \midrule
    Length-only (3 vs 3 turns) & $\approx 0.50$ & --- & --- \\
    CRA-convex (full) & 0.0826 & --- & 0.9769 \\
    CRA-Net ($\lambda=0$, no length reg.) & 1.0000 & [1.0000, 1.0000] & 1.0000$^{\dagger}$ \\
    CRA-Net ($\lambda=0.05$, + length reg.) & 0.9917 & [0.9865, 0.9960] & 0.4545 \\
    \bottomrule
  \end{tabularx}
\end{table*}

\subsection{Ablation analysis}
\label{sec:cosafe-ablation}

Removing $S_1$ collapses discrimination: the score fails to separate CRA-positive
from CRA-negative sessions and yields sFPR $=1.0$. Removing $S_2$ preserves AUROC
but delays warning time (TTD increases by approximately 0.4 turns), consistent
with $S_2$ contributing early-turn evidence. Removing $S_3$ has minimal effect in
this dataset under the lightweight proxy.

\subsection{Diagnostic summary}
\label{sec:cosafe-summary}

CoSafe demonstrates that trajectory-level scores \emph{can} separate gradual
escalation from short direct attacks in a large public corpus, but the split is
length-confounded and negatives are not benign.

Exploratory probe on WildChat-1M~\citep{zhao2024wildchat}: after streaming
200{,}000 English sessions we identified only N = 59 moderation-defined
escalating positives and 59 matched benign negatives. Length-only AUROC was
0.68; CRA-convex 0.59; CRA-Net (CoSafe$\to$WildChat) reported AUROC 1.00 with
theta = 1.0 saturation, so bootstrap CIs and sFPR are unreliable at this sample
size. We do not cite WildChat as a transfer result; Human-CRA-Transfer
(Section~\ref{sec:human-transfer}) is the human-authored validation set.
Primary scientific claims rest on
CRA-Bench (Section~\ref{sec:primary-eval}), mixed-split calibration
(Section~\ref{sec:primary-benign}), and evasion stress tests
(Section~\ref{sec:primary-evasion}).

\section{Architectural Composability and Deployment Patterns}
\label{sec:deployment}

A central design goal of the CRA Framework is composability: organizations
\emph{can} add a session-layer monitor alongside existing turn-level
guardrails. CRA consumes the same per-turn inputs (user message, model
response, and optionally tool traces) and produces an independent session-level
signal that can be combined with existing allow/deny decisions. The benign-FPR
numbers in this paper use ShareGPT and other \textbf{public-chat proxies}; we
have not yet validated FPR on customer RAG logs, AgentBench traces, or classroom
tutoring corpora.

\begin{figure}[!t]
  \centering
  \includegraphics[width=0.95\columnwidth]{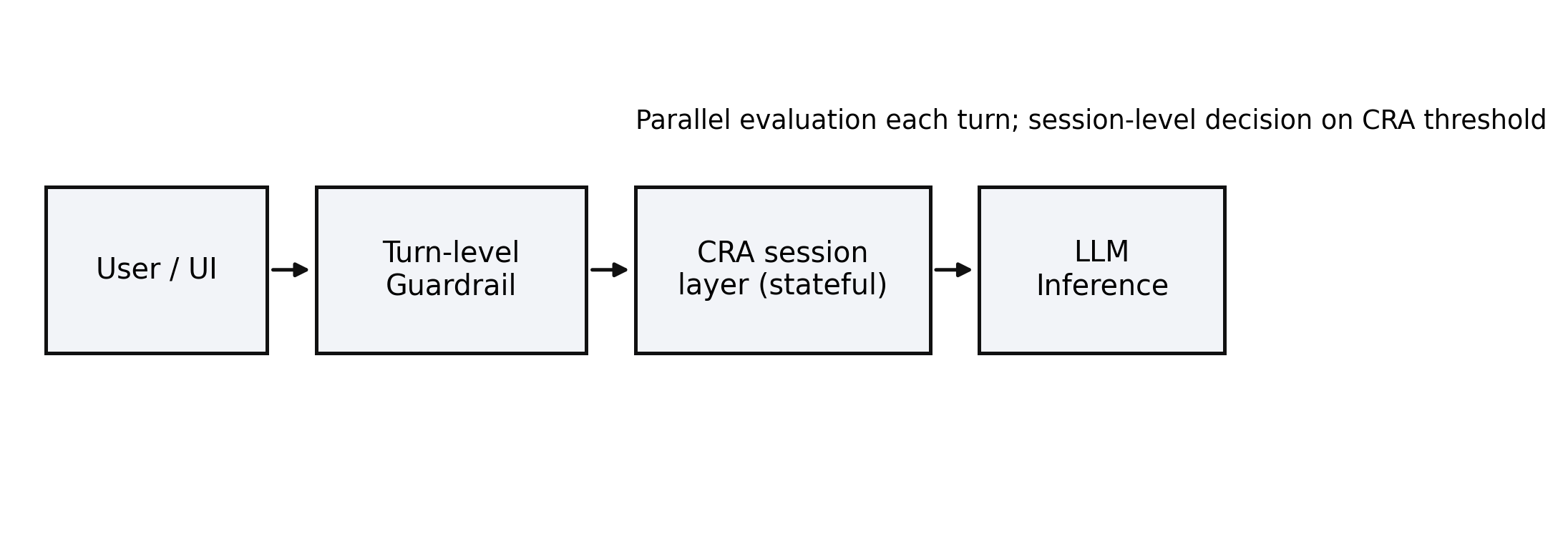}
  \caption{Reference deployment pattern: CRA runs as a stateful session layer
  alongside existing turn-level controls. Both layers observe the same user inputs
  and model outputs; turn-level policy checks remain unchanged, while CRA adds a
  session-level intervention signal.}
  \label{fig:make-arch}
\end{figure}

\paragraph{Scope note.}
The deployment patterns below are architectural sketches intended to guide
implementation and evaluation. They are not presented as validated case studies
and should be treated as hypotheses to be tested with domain-grounded telemetry
and red-team sessions. Figure~\ref{fig:make-arch} shows a reference integration.

\subsection{Integration with Enterprise Moderation and Policy Pipelines}
\label{sec:enterprise_integration}

In enterprise settings, CRA is intended to sit alongside existing enforcement
layers (prompt/response moderation, PII detectors, and allow/deny tool policies).
A practical integration pattern is to emit a structured event per turn containing
$\mathrm{CRA}(t)$, the top contributing sub-signals, and a stable policy/version
identifier into the same logging and audit stream used for moderation. This
supports both (i)~online actions such as interventions and routing, and
(ii)~offline governance workflows including case review, compliance reporting,
and post-incident reconstruction.

\subsection{Enterprise RAG Deployment}

In enterprise retrieval-augmented generation (RAG) systems, the Information
Accumulation Graph (IAG) can be grounded in artifacts the organization already
maintains. Many enterprises classify documents, entities, and individual fields
into sensitivity tiers (e.g., public, internal, confidential, restricted). These
tiers can initialize IAG node weights directly, rather than relying only on what
the model discloses in a given response. Concretely, if a retrieved record is
tagged \emph{confidential}, entities and attributes extracted from it can inherit
a higher base weight so that repeated exposure about the same entity drives $S_2$
upward earlier.

RAG also provides a natural reference frame for semantic drift. Instead of
anchoring drift to a user-declared intent (often missing or vague in enterprise
chat), the drift monitor can be anchored to the retrieved context itself. A
practical approach is to maintain an embedding centroid over retrieved chunks
from the last $k$ turns and score drift as distance from that centroid. This
flags sessions where the dialogue moves beyond retrieved, grounded context, which
can correlate with out-of-scope assistance and hallucination.

Finally, CRA fits naturally into cost- and risk-aware routing. Many production
RAG stacks route queries across models, tools, or retrieval strategies based on
cost and policy constraints. A session-level CRA score can act as an additional
routing signal: as CRA rises, the system can tighten retrieval (e.g., restrict to
lower-sensitivity corpora), require stronger authentication for sensitive
namespaces, switch to a more conservative model, and/or introduce friction such
as confirmations and audit logging.

\subsection{Agentic Pipeline Deployment}

Agentic LLM systems introduce a CRA attack surface that is weaker or absent in
pure chat: tool use. An agent can execute web requests, database queries, code
execution, file operations, and business workflow APIs. Each action may be
authorized in isolation, yet the sequence can produce an outcome that was never
intended to be authorized, for example enumerating records through many small
queries or chaining tools to extract sensitive data.

The CRA Framework extends to agents by treating tool calls and their outcomes as
first-class turns in the session history. In this setting, the IAG is updated not
only from language output, but also from structured traces: which tables and
fields were queried, which files were read or written, which endpoints were
called, and which identifiers were touched. \emph{Entities} include operational
objects such as accounts, tickets, repositories, and cloud resources;
\emph{attributes} include accessed fields, permissions, and derived artifacts.

The compliance-gradient detector can also be reframed for agents. Rather than
tracking refusal language alone, it can track escalation in capability and
resource access: trends in tool privilege, the breadth of data sources touched,
the rate of sensitive reads, and the fraction of actions requiring elevated
scopes. A sustained increase in these quantities, even without explicitly unsafe
language, can indicate that an agent is being guided toward higher-risk behavior.

\subsection{Educational AI Deployment}

Educational AI systems present a qualitatively different deployment pattern: the
primary CRA risk is often a gradual slide in pedagogical scope rather than a
direct adversarial attack. A student may begin by requesting hints, then ask for
increasingly specific guidance, and eventually obtain full solutions or
near-verbatim answers. Each individual request can appear reasonable in
isolation, but the session as a whole can outsource the cognitive work the
assignment was designed to elicit.

In this setting, the semantic drift monitor can be anchored to the declared
learning objective, rubric, or allowed-help policy (e.g., ``explain the concept,''
``give a hint,'' ``check my work; do not solve''). Drift then measures movement
from supported scaffolding toward unsupported substitution. The
compliance-gradient component can be instantiated as a shift in response style
(conceptual explanation $\to$ step-by-step completion $\to$ answer-giving) using
features such as response specificity, directness, and the presence of final
answers.

Interventions in this context should therefore look different from security
guardrails. Instead of hard blocking, the system can steer: ask the student to
show an attempt, offer a hint template, provide concept-level feedback, or switch
into a Socratic question-asking mode. In this form, CRA functions as a
pedagogical integrity guardrail enabled by session-level accumulation signals.

\FloatBarrier
\section{Open Problems and Research Agenda}
\label{sec:open_problems}

The CRA Framework is intended as a practical scaffold for session-layer
monitoring, but several research problems must be addressed for robust
deployment and evaluation.

\subsection{IAG Maintenance at Scale}

The Information Accumulation Graph is the most structurally novel component of
the framework, but it can also be the most computationally demanding. In long
sessions with unbounded entity growth (common in agentic workloads) maintaining a
full IAG may be intractable. Sliding-window and bounded-memory approximations
improve tractability, but can underestimate accumulation when entities reappear
after long gaps: an entity introduced at turn 5 and revisited at turn 45 may be
missed by a short window. Developing efficient approximate IAG data structures
with bounded error, along with principled pruning and decay strategies, remains an
open problem.

\subsection{Threshold Calibration Without Ground Truth}

Selecting $\theta$ (and related soft-warning thresholds) requires characterizing
the CRA score distribution under both benign and adversarial sessions. Benign data
is typically abundant in production logs, whereas CRA-positive sessions are rare
and often unlabeled, yielding an imbalanced, partially observed setting. Practical
calibration methods must therefore combine weak supervision (rubrics, red-team
generation, and heuristics), uncertainty-aware tuning, and robustness to
distribution shift across model versions, policies, and user populations.

\subsection{Explainable Intervention}

When CRA triggers a session-level intervention, the system should provide
actionable feedback without disclosing detection details that enable adaptive
evasion. The proposed decision certificate is one approach: a structured artifact
that reports which signals increased, summarizes the relevant session fragments at
a policy-safe level, and supports auditing. Formalizing what should be shown to
end users versus operators, and how to standardize certificates for compliance
workflows, is both a technical and a governance challenge.

\subsection{Adversarial Robustness of Sub-Signals}

A capable adversary may attempt to keep drift ($S_1$), accumulation ($S_2$), and
compliance trend ($S_3$) simultaneously low: for example, by maintaining topical
similarity while extracting sensitive facts, spreading disclosures across many
entities, or mixing refusals to flatten slopes. Understanding the hardness of
joint evasion, and quantifying whether heterogeneous fusion meaningfully raises
attacker cost relative to single-signal detectors, is an open security question
with direct implications for guardrail design.

\FloatBarrier
\section{Discussion and Limitations}
\label{sec:discussion}

The illustrative results in Section~\ref{sec:illustrative} show that the CRA
Framework produces distinct, theoretically consistent signal trajectories across
multiple CRA threat types, supporting the claim that multi-turn risk can remain
locally invisible at the turn level while becoming detectable as a session-level
process. The CoSafe evaluation in Section~\ref{sec:experiments} extends this
finding to real multi-turn logs, but several limitations are important for
interpretation.

\paragraph{Heuristic instrumentation.}
The reference implementation uses lightweight, reproducible proxies. $S_2$ relies
on simple entity tracking and approximate coverage signals rather than a
domain-specific sensitivity taxonomy or calibrated PII detectors, and $S_3$ uses
refusal and hedging cues rather than a trained compliance classifier. These
choices reduce engineering dependencies, but the resulting scores should be
interpreted as relative indicators of accumulation dynamics rather than
calibrated probabilities of harm.

\paragraph{Dataset structure and signal specificity.}
The CoSafe evaluation has a structural length confound (all CRA-positive sessions:
3 turns; all CRA-negative sessions: 1 turn). Length-only classification achieves
AUROC $= 1.0000$, meaning CRA's convex-weight headline AUROC does not demonstrate
content-based discrimination on its own. The length-matched evaluation
(Section~\ref{sec:length-matched}) quantifies this gap: convex-weight CRA achieves
AUROC $= 0.083$ against benign 3-turn ShareGPT conversations (below chance),
because $S_1$ (semantic drift) responds to topical diversity rather than harmful
content specifically. The benign-FPR evaluation (Section~\ref{sec:benign-fpr})
shows 100\% of benign multi-turn conversations exceed the CoSafe-calibrated
threshold under the proxy signals.

The key finding is that the choice of \emph{fusion strategy} matters critically:
CRA-Net with gradient-reversal length regularization ($\lambda=0.05$) achieves
AUROC $= 0.9917$ on the same length-matched corpus, with sFPR $= 0.4545$. The
learned trajectory model, forced to suppress the length shortcut, discovers
content-based patterns that convex-weight fusion cannot represent. This result
motivates CRA-Net as the deployment-ready fusion strategy and identifies
improving the proxy signals ($S_2^{\mathrm{ref}}$, keyword $S_3$) as the primary
path to reducing benign false alarms.

\paragraph{Interventions are deployment-specific.}
CRA identifies when a session is trending toward a policy-relevant unsafe state,
but it does not prescribe what action to take. Appropriate interventions depend on
application constraints: customer support may prefer low-friction nudges and
re-anchoring; enterprise assistants may require step-up authentication, audit
logging, or human review; and agentic systems may need to pause or sandbox tool
access, prune untrusted context, or switch to a conservative safe mode. We
therefore treat the CRA score and decision certificate
(Section~\ref{sec:certificates}) as inputs to a broader governance and response workflow, rather than a standalone block/allow mechanism.

\FloatBarrier
\section{Conclusions}
\label{sec:conclusions}

The assumption that safety is a property of individual exchanges is rarely a
conscious design commitment; it is more often an inherited simplification.
Turn-level guardrails are attractive because they are easy to instrument,
benchmark, and scale: each prompt--response pair can be scored in isolation.
However, as LLM deployments mature toward long-horizon conversational agents,
agentic pipelines with tool use, and enterprise systems that handle sensitive
information across extended sessions, this stateless view becomes a structural
liability. In these settings, harm is frequently a trajectory property: it emerges
from drift, accumulation, and gradual boundary erosion that is not visible in any
single message.

In response, we formalized Conversational Risk Accumulation (CRA) as a class of
multi-turn safety failures that are, by construction, invisible to purely stateless
guardrail functions. We then introduced the CRA Framework, a session-layer
guardrail architecture designed to complement (not replace) existing turn-level
controls. The framework operationalizes three signals corresponding to distinct
accumulation mechanisms: (i)~a Semantic Drift Monitor to detect displacement from
an anchored intent or scope; (ii)~an Information Accumulation Graph to represent
cross-turn disclosure as a structured, growing knowledge state; and (iii)~a
Compliance Gradient Detector to capture gradual changes in refusal and hedging
behavior. Fused into a session-level CRA score, these signals enable interventions
triggered by the evolving conversation state rather than by any single exchange.

Empirical grounding comes from \textbf{CRA-Bench v0.1} and the
LLM-paraphrased \textbf{CRA-Bench v0.2} under the same primary protocol
(Section~\ref{sec:primary-eval}, Section~\ref{sec:bench-v02}): 1{,}200
same-length sessions per release (600 positive, 600 topic-matched
benign-twin) across three threat families, session-level splits, and
mixed validation calibration. On v0.1, convex and turn-only baselines sit
near chance (AUROC $\in [0.54, 0.60]$, sFPR $\geq 0.67$ at TPR$=0.90$);
vanilla CRA-Net ($\lambda{=}0.05$) attains AUROC $0.993$ with sFPR
$0.008$; a Judge-LLM transcript-level ceiling reaches AUROC $1.000$ but
is not deployable. On the paraphrased v0.2 stress test, vanilla CRA-Net
falls by $0.16$ AUROC to $0.829$ and \emph{inverts} on paraphrased
fragmentation (AUROC $0.42$), exposing the
\emph{per-method template-memorization budget} between v0.1 and v0.2.

We introduce \textbf{CRA-Net DA} (Section~\ref{sec:cranet-da}): a
domain-adversarial variant that adds a gradient-reversal head predicting
the threat family from the encoder's session representation. At 3
families, CRA-Net DA holds AUROC at $0.919$ across v0.1 and v0.2
($\Delta=0$ under paraphrase) and cuts v0.2 sFPR from $0.425$ to
$\mathbf{0.175}$ at $\lambda_{\mathrm{fam}}{=}0.3$
(Tables~\ref{tab:primary-main}, \ref{tab:primary-v01-vs-v02}). Extending
the bench to \textbf{5 families} (adding \textit{persona priming} and
\textit{context stuffing}; Section~\ref{sec:bench-5fam},
Tables~\ref{tab:primary-5fam}, \ref{tab:lofo-scaling}) yields two main
findings. First, \textbf{mean LOFO diagnostic AUROC scales with training
family count} ($0.384 \to 0.628$ vanilla; $0.387 \to 0.647$ CRA-Net DA at
$\lambda_{\mathrm{fam}}{=}0.1$), but held-out fragmentation and aggregation
often remain below chance even at four training families---so we position CRA-Net
as a \textbf{within-distribution session scorer} calibrated on CRA-Bench and
Human-CRA-Transfer, not as a cross-family universal guardrail. Second, the right deployment-time calibration is
\textbf{benign-anchored}: pick $\theta$ to control benign FPR on
external chat (we use $N{=}1{,}000$ multi-turn ShareGPT sessions), then
report TPR on CRA-Bench. At a $1\%$ FPR budget,
\textbf{CRA-Net DA reaches TPR $\mathbf{=1.000}$ on the CRA-Bench v0.2
test split} ($\theta{=}0.492$); vanilla CRA-Net at the same budget
reaches TPR $0.640$ ($\theta{=}0.774$).
Table~\ref{tab:benign-fpr-5fam} is the new headline operating point and
supersedes the fixed-TPR row of Table~\ref{tab:benign-fpr} as the
deployment-time recommendation. Honest qualification: even at 4
training families, held-out fragmentation and aggregation remain below
$0.5$ AUROC, and $\lambda_{\mathrm{fam}}$ must be re-tuned per family
pool size ($0.3$ at $N{=}3$, $0.1$ at $N{=}5$). \textbf{Human-CRA-Transfer} (Section~\ref{sec:human-transfer},
Table~\ref{tab:human-transfer}) closes the synthetic$\to$real gap:
CRA-Net trained only on CRA-Bench reaches AUROC $0.919$ on 750
human-authored CoSafe gradual-escalation sessions, while Llama Guard~3
falls to $0.695$ on the same corpus despite near-perfect synthetic-bench
scores---highlighting why human transfer subsets matter alongside
in-distribution guardrail ceilings. CoSafe diagnostics
(Section~\ref{sec:cosafe-diagnostic}) and benign-traffic calibration
(Section~\ref{sec:benign-fpr}) complement the primary CRA-Bench and
Human-CRA-Transfer evidence; the small-$N$ WildChat probe is footnoted only.

Several open problems remain before stateful guardrails become routine
production infrastructure. The most immediate is \textbf{held-out-family
coverage on CRA-Bench}: even at 4 training families, LOFO diagnostics show
fragmentation and aggregation AUROC below $0.5$, so cross-template deployment
without retraining is not yet supported. Expanding to 7--10 mechanistically
distinct families (or accepting a narrower within-distribution deployment scope)
is the plausible path; adversarial regularization alone has a narrow
$\lambda_{\mathrm{fam}}$ window that contracts as the family pool grows. Efficient IAG maintenance is required for long
sessions and agentic traces with many entities and tool outputs;
threshold and weight calibration must be robust under distribution shift
and adversarial adaptation (the benign-anchored recipe in
Table~\ref{tab:benign-fpr-5fam} is a strong default but does not preclude
adaptive attack against the ShareGPT-like calibration distribution); and
interventions must be explainable to users and auditable by operators
without disclosing exploitable detection logic. A key security question
is whether composite, multi-signal monitoring with benign-anchored
calibration meaningfully raises the difficulty of evasion compared to
isolated turn-level filters.

The field has made substantial progress in improving the safety of individual LLM
outputs. The next necessary investment is to make conversations and the systems
built on top of them safe by design.

\FloatBarrier
\section*{Reproducibility}
\label{sec:reproducibility}
To support replication of the experiments in Section~\ref{sec:experiments} and Table~\ref{tab:cosafe-main}, we release the following artefacts under permissive licences:
\begin{itemize}
  \item CRA reference implementation (all fusion strategies): convex fusion (Section~\ref{sec:convex_fusion}), CRA-Net (Section~\ref{sec:cranet}), \textbf{CRA-Net DA} (Section~\ref{sec:cranet-da}, the family-adversarial variant exposed in \texttt{CRANetDA} and \texttt{train\_cranet\_da} inside \path{run_cranet.py} and \path{run_cra_extended_protocol.py}), \textbf{Deep CORAL alignment} (\texttt{train\_cranet\_coral}, \texttt{coral\_loss} in \path{run_cranet.py}), feature extraction (all-MiniLM-L6-v2, spaCy NER, $S_3$ lexicon), \path{generate_cra_bench_v01.py}, \path{run_cra_primary_protocol.py}, \path{run_cra_extended_protocol.py} (primary Tables~\ref{tab:primary-main}, \ref{tab:primary-perfamily}, \ref{tab:primary-lofo}, \ref{tab:primary-v01-vs-v02}, \ref{tab:benign-fpr}, \ref{tab:primary-5fam}, \ref{tab:lofo-scaling}, \ref{tab:benign-fpr-5fam}, \ref{tab:human-transfer}; results JSONs \path{cra_extended_protocol_v0_1_da.json}, \path{cra_extended_protocol_v0_2_da_full.json}, \path{cra_extended_protocol_v0_2_5fam_final.json}, \path{cra_extended_protocol_v0_2_5fam_coral.json}, and \path{cra_extended_protocol_v0_2_3fam_coral.json}), the GPT-4o-mini judge baseline (\path{run_judge_cra_bench.py}), guardrail baselines (\path{run_llamaguard_cra_bench.py}; Llama Guard~3 via Ollama \path{llama-guard3:1b} and Qwen3Guard; results \path{llamaguard_llama_guard3_1b_v02_5fam.json}, \path{llamaguard_llama_guard3_human_transfer.json}, \path{llamaguard_qwen3guard_human_transfer.json}), synthetic-to-real transfer (\path{run_human_transfer_eval.py}; \path{human_transfer_eval.json}), and diagnostic CoSafe harnesses (Table~\ref{tab:cosafe-main}).
  \item CRA-Bench v0.1 (Section~\ref{sec:cra_bench}), CRA-Bench v0.2 (Section~\ref{sec:bench-v02}), the \textbf{5-family CRA-Bench v0.2} expansion (Section~\ref{sec:bench-5fam}), and \textbf{Human-CRA-Transfer} (Section~\ref{sec:human-transfer}): the generated JSONL sessions, the generation scripts (with fixed seeds and a pinned model snapshot), the v0.2 LLM paraphrase cache (\path{paraphrases_v0_2.json}, produced by \path{build_paraphrases_v0_2.py} with \path{LLM_MODEL=gpt-4o-mini}, now covering 139 templates across all five families), and the template specifications for each of the five CRA threat types including the two new families \emph{persona priming} and \emph{context stuffing}.
  \item Figure-generation scripts and the synthetic CRA timeseries (\path{cra_timeseries.csv}) used to produce the illustrative figures in Section~\ref{sec:illustrative}.
  \item Latency, convex grid-search, and $S_3$ validation:
  \path{run_latency_benchmark.py}, \path{run_convex_grid_search.py},
  \path{run_s3_proxy_validation.py} (Appendix~\ref{sec:appendix-paraphrase}--\ref{sec:appendix-s3}).
\end{itemize}

The release tag, commit hash, and SHA-256 of the CRA-Bench v0.1 JSONL bundle are listed in the repository README so that replications can pin exact artefacts. Production conversation logs, where applicable, are retained under organisational policy and are not published.

\section*{Data Availability}
The CoSafe benchmark~\citep{yu2024cosafe} and public rollout corpus~\citep{cosafe2024} are available at \url{https://huggingface.co/datasets/Asap7772/cosafe_all_rollouts}. CRA-Bench v0.1 and the reference implementations of both CRA fusion strategies (see Reproducibility statement above) will be released at the camera-ready stage at a permissive-licence public repository (URL withheld for double-blind review; an anonymized snapshot is provided to reviewers via the supplementary material). Raw production conversation logs, where applicable, are retained under organisational policy and are not published.

\section*{Conflict of Interest}
The authors declare no conflicts of interest.

\section*{Institutional Review}
Not applicable: this manuscript describes a technical framework and illustrative synthetic traces without human subjects data.

\section*{Appendix: Reproducibility Artefacts}
\subsection{Paraphrase prompt (CRA-Bench v0.2)}
\label{sec:appendix-paraphrase}
The v0.2 paraphrase cache is produced by \url{build_paraphrases_v0_2.py} using
\url{LLM_MODEL=gpt-4o-mini}, \url{temperature=0.8}, and the prompt template
\url{PARAPHRASE_PROMPT} in that file (eight variants per template, placeholders
preserved verbatim). The full JSON cache is
\url{data/cra_bench_v01/paraphrases_v0_2.json}.

\subsection{$S_3$ proxy validation protocol}
\label{sec:appendix-s3}
\texttt{run\_s3\_proxy\_validation.py} samples 200 user-window snippets from
CRA-Bench v0.2, scores the keyword $S_3$ proxy (threshold $0.5$), and compares
binary unsafe labels from Llama~Guard~3 (Ollama \texttt{llama-guard3:1b}) on the
same prefix. On our run ($n{=}200$ scored, $0$ Ollama skips): Cohen's
$\kappa{=}0.00$, raw agreement $11.5\%$, so the keyword proxy is \emph{not}
interchangeable with guard labels ($\kappa \geq 0.7$ was our bar for replacing
it in production). We therefore keep the keyword slope as the reproducible
default in all tables while documenting this audit for future DeBERTa-$S_3$
calibration.

\subsection{Worked decision certificate}
\label{sec:appendix-cert}
\begin{lstlisting}[basicstyle=\ttfamily\footnotesize,breaklines=true]
{
  "session_id": "bench_fragmentation_pos_0042",
  "policy_version": "cra-enterprise-v1.2",
  "cra_score": 0.61,
  "threshold_band": "soft_warn",
  "top_signals": [
    {"name": "S1_drift", "contribution": 0.28},
    {"name": "S2_iag", "contribution": 0.22},
    {"name": "S3_refusal_slope", "contribution": 0.11}
  ],
  "iag_delta": {"new_entities": ["Project Orion"], "tier": "confidential"},
  "drift_trend": "rising since turn 4",
  "refusal_trend": "declining over last 3 assistant turns",
  "operator_summary": "Session intent drifted from benign planning toward
    sensitive operational detail; cumulative disclosures crossed internal tier."
}
\end{lstlisting}

\bibliographystyle{unsrtnat}
\bibliography{references}

\end{document}